\documentclass[final,5p,times,twocolumn,authoryear]{elsarticle}
\usepackage{algorithm}
\usepackage{algorithmic}
\usepackage{amsfonts}
\usepackage{amsmath}
\usepackage{amssymb}
\usepackage{array}
\usepackage{booktabs} % For even nicer tables.
\usepackage{caption}
\usepackage{color}
\usepackage{epstopdf} % package to include eps. files
\usepackage{float}
\usepackage{graphicx}
\usepackage{multirow} % Multirow is for tables with multiple rows within one cell.
\usepackage{stfloats}
\usepackage{subfig}
\usepackage{textcomp}
\usepackage{url}
\usepackage{verbatim}
\usepackage{xcolor}
\usepackage{subfiles} % load a tex file as a chapter, Best loaded last in the preamble
\usepackage[hidelinks]{hyperref} % for hyperlinks
\definecolor{darkred}{RGB}{170, 0, 0}
\newcommand{\modified}[1]{#1}

\journal{Arxiv}

\begin{document}

\begin{frontmatter}

\title{Unveiling Stochasticity: Universal Multi-modal Probabilistic Modeling for Traffic Forecasting}

\author[epfl]{Weijiang Xiong}
\ead{weijiang.xiong@epfl.ch}

\author[epfl]{Robert Fonod}
\ead{robert.fonod@ieee.org}

\author[epfl]{Nikolas Geroliminis\corref{cor1}}
\ead{nikolas.geroliminis@epfl.ch}
\cortext[cor1]{Corresponding author.}

\affiliation[epfl]{organization={Urban Transport Systems Laboratory (LUTS), EPFL},
            city={Lausanne},
            postcode={1015},
            country={Switzerland}}

\begin{abstract}
Traffic forecasting is a challenging spatio-temporal modeling task and a critical component of urban transportation management.
Current studies mainly focus on deterministic predictions, with limited consideration of the uncertainty and stochasticity in traffic dynamics.
Therefore, this paper proposes an elegant yet universal approach that transforms existing models into probabilistic predictors by replacing only the final output layer with a Gaussian Mixture Model (GMM) layer, \modified{a network module inspired by the Mixture Density Network (MDN).}
The modified model requires no changes to \modified{data preparation, optimizer or hyperparameters}, and can be trained using only the Negative Log-Likelihood (NLL) loss, without any auxiliary or regularization terms.
Experiments on multiple traffic datasets show that our approach generalizes from classic to modern model architectures.
\modified{We propose a systematic evaluation procedure and demonstrate that our approach consistently improves CRPS over unimodal and deterministic baselines, and generates more informative and better-calibrated predictive intervals for a wide range of scenarios.
Meanwhile, GMM adaptation often improves L2 error with trade-offs in L1 error metrics.}
Finally, a more detailed study on a real-world dense urban traffic network is presented to examine the impact of data quality on uncertainty quantification and to show the robustness of our approach under imperfect data conditions.
\modified{Code available at \href{https://github.com/Weijiang-Xiong/OpenSkyTraffic}{https://github.com/Weijiang-Xiong/OpenSkyTraffic}}
\end{abstract}


\end{frontmatter}

\section{Introduction}
\label{sec: introduction}
Spatio-temporal modeling is a fundamental task in machine learning with applications in various domains, such as sustainable power systems \citep{bessa2014spatial}, climate change \citep{maclean2020predicting}, transportation systems \citep{yuan2021survey} and video data analysis \citep{zhou2025leveraging}.
Specifically, traffic forecasting has been an active research topic in this domain, owing to its vital role in smart cities \citep{Tedjopurnomo2022survey, chen2019review}.
With a traffic forecasting model, future traffic states of the city can be predicted using historical data, providing essential grounds for adaptive traffic management \citep{ravish2021intelligent, wang2018review}.
In earlier research, statistical methods, such as Historical Average (HA) and Autoregressive Integrated Moving Average (ARIMA), were developed as baseline predictors that use the periodic patterns and temporal dynamics of traffic \citep{vlahogianni2014short}.
However, these approaches have limited performance because traffic data contain complex dependencies in both spatial and temporal dimensions: the future state of one road depends on its own dynamics and on nearby roads \citep{ermagun2018spatiotemporal}.
As a result, the modern research trend focuses on learning these correlations to improve prediction accuracy \citep{jiang2022graph}.
Since the collected data are deterministic, i.e., only one value is measured for a traffic state variable at a specific location and time, the most natural evaluation is to compare error metrics, such as Mean Absolute Error (MAE) \citep{ferreira2023forecasting}.

\begin{figure}[ht]
    \centering
    \includegraphics[width=0.8\linewidth]{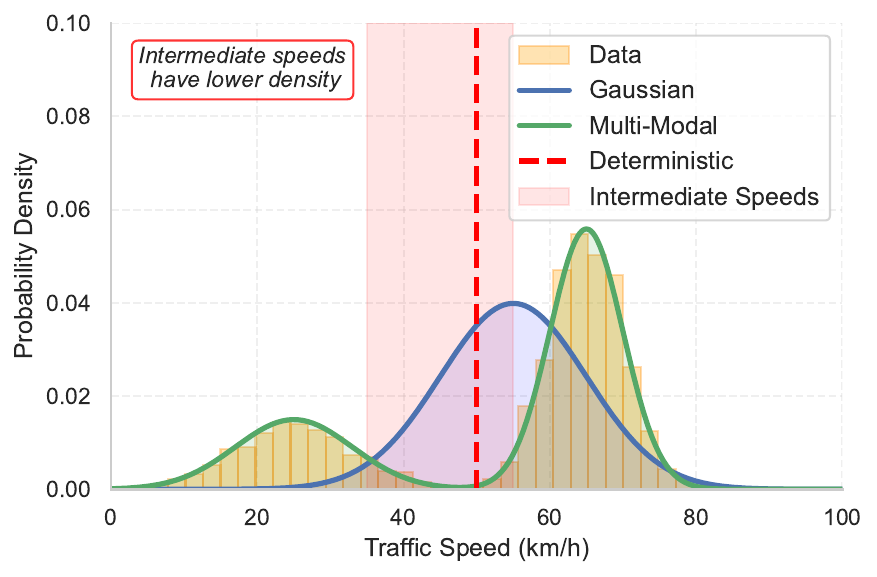}
    \caption{Multi-modal data distribution and various predictions}
    \label{fig:data_distribution_and_predictions}
\end{figure}

These facts and practices have shaped the paradigm for developing and evaluating traffic forecasting methods, and traffic forecasting has long been framed in a deterministic setting, i.e., one future value is predicted and evaluated.
However, urban traffic is highly dynamic and uncertain due to various factors, such as traffic signals, pedestrian activities, weather conditions, traffic incidents, and the randomness in drivers' behaviors \citep{mauro2015traffic}.
As a result, the process that generates deterministic traffic data is stochastic, and the collected data can be regarded as one realization of that process.
From the perspective of modeling, the future traffic variables should be modeled with probabilistic distributions instead of deterministic values \citep{ghosh2010random}.
On the other hand, real-world decision-making processes, such as traffic control, should consider all possibilities to provide safe and robust solutions covering all outcomes \citep{wanke2004modeling}.
Moreover, the randomness in traffic can have a big impact and result in distinct outcomes.
For example, travel time on an arterial road can have multiple possible values due to uncertain waiting times at traffic signals \citep{ramezani2012travel}.
Consequently, the data distribution can be multi-modal, i.e., having multiple peaks in the probability density function.

Figure~\ref{fig:data_distribution_and_predictions} illustrates different predictive methods for a multi-modal traffic speed distribution, where a road segment can be either congested or in free flow and intermediate states are less common.
This scenario is realistic when the traffic is highly uncertain (especially for urban traffic) or when existing information is insufficient to distinguish the two states (e.g., for long-term predictions).
In such a case, predicting a deterministic estimate can conceal the uncertainty and lead to false confidence, even though it can have low deterministic error metrics when averaged over multiple trials.
A unimodal distribution (e.g., Gaussian) can indicate the level of uncertainty with a large variance; however, this prediction can be too vague to provide useful information.

Building on these insights, this paper proposes to model traffic variables as multi-modal random variables. 
We create a universal and effective approach to enhance traffic forecasting models by expanding the prediction scope from a most-likely value to multi-modal probabilistic distributions.
We show that the adaptation requires only a minimal modification: replacing the last output layer with a Gaussian Mixture Model (GMM) layer, yet the performance gain and robustness are consistently demonstrated through experiments on various datasets.
With the expressive power of multi-modal predictive distributions, this approach can unveil the underlying stochasticity in deterministic data and provide a stronger basis for decision-making processes that rely on these predictions, e.g., traffic signal control in transportation systems \citep{tsitsokas2023two}.
Despite the simplicity of the model modification, improvements in probabilistic evaluation metrics are significant, while deterministic prediction performance remains comparable.

In summary, the main contributions of this paper are:
\begin{itemize}
    \item A simple yet effective approach to enhance the descriptive power of traffic forecasting models by expanding the prediction space into multi-modal probabilistic distributions.
    \item A general and comprehensive evaluation framework for multi-modal probabilistic predictors, applicable whenever probability density is available.
    \item Extensive experiments on multiple traffic forecasting datasets to demonstrate the advantages of multi-modal probabilistic predictions.
    \item A case study to show the impact of data quality on uncertainty and highlight the robustness of our approach.
\end{itemize}

\section{Related Work}
\label{sec: related work}
This section provides a brief review of the two most related research topics, i.e., traffic forecasting and uncertainty quantification, since our approach lies at their intersection.

\subsection{Traffic Forecasting}
\label{subsec: traffic forecasting}
Traffic forecasting has been a widely studied research topic due to its significance both as a complex spatio-temporal modeling problem and as a practical application in transportation systems \citep{vlahogianni2014short}.
Most modern approaches are based on machine learning, where the models are designed to cope with the dependencies in both spatial and temporal dimensions \citep{liu2019tailored, jiang2022graph}.
The Diffusion Convolutional Recurrent Neural Network (DCRNN) is a pioneering work that learns the spatial relations through a diffusion convolution over the graph structure of the transportation network and captures the temporal correlations using a Recurrent Neural Network (RNN) \citep{li2018dcrnn}.
To improve the efficiency of temporal modeling, the Spatio-Temporal Graph Convolutional Networks (STGCN) applies a convolution over the time dimension, thus avoiding the sequential nature of RNNs \citep{yu2018STGCN}.

In fact, traffic conditions at different locations are often intertwined, and data correlations often go beyond geographical distances \citep{kwak2021traffic, li2021multistep}.
Therefore, many follow-up works have focused on adaptively learning relations from data.
The Graph WaveNet (GWNet) \citep{wu2019graph} proposes enhancing the pre-defined graph with an adaptive graph, which is constructed by the pairwise product of learnable node embeddings.
Other practices include by incorporating various attributes \citep{xiao2026heuristic} and building dynamic graph convolution \citep{zhang2026adaptive}.
Specifically, attention mechanism is an effective approach for modeling complex spatio-temporal correlations, and have been applied in various approaches~\cite{do2019effective, liu2025multilayer}.
The Attention-based Spatial-Temporal Graph Convolutional Networks (ASTGCN) \citep{guo2019attention} further introduces a more dynamic attention mechanism \citep{vaswani2017attention} for spatial and temporal dimensions, which effectively assigns weights to the spatial or temporal neighbors according to the learned correlation patterns.
Similarly, the Graph Multi-Attention Network (GMAN) \citep{zheng2020gman} employs an encoder-decoder structure with an attention mechanism on both spatial and temporal dimensions.
In addition, the encoder messages are enhanced by a transform attention between past and future time steps, allowing the model to alleviate error propagation.

The feature embedding technique also plays an important role in traffic forecasting.
The Spatial and Temporal Identity Information (STID) \citep{shao2022spatial} identifies that a bottleneck in modern traffic forecasting models is the lack of sufficient information to distinguish different samples, and proposes augmenting feature embeddings with metadata, such as time of day and day of week.
The Multivariate Time Series Graph Neural Network (MTGNN) \citep{wu2020connecting} designs a graph learning module that combines the node embeddings with static features to formulate more comprehensive node descriptions.
Additionally, MTGNN organizes message exchanges for neighbors at different distances, enabling larger receptive fields compared to the traditional GNNs.
A similar idea has also been utilized in the Dynamic Multi-hop Graph Attention Network (DMGAN) \citep{li2022dmgan}.
Later, a Spatio-Temporal Adaptive Embedding (STAE) technique has been introduced to augment the traffic value embeddings with a set of learnable adaptive embeddings \citep{liu2023spatio}.
As a result, the STAE technique can significantly improve the performance of a vanilla Transformer \citep{vaswani2017attention}.
\modified{
A recent state-of-the-art method, Hierarchical U-Net Transformer (HUTFormer) \citep{shao2025hutformer}, further combines an efficient input embedding strategy with multi-scale representation of traffic data, and it has demonstrated favorable performance in long-term traffic forecasting.
}

\subsection{Uncertainty Quantification}
\label{subsec: uncertainty quantification}

Research on uncertainty quantification falls within the context of Bayesian Deep Learning, where uncertainty is generally categorized as aleatoric and epistemic uncertainties \citep{kendall2017uncertainties}.
Epistemic uncertainty reflects insufficient knowledge to determine the model parameters, and can be reduced by a larger collection of training data.
In contrast, aleatoric uncertainty refers to the inherent stochasticity in data, which originates from the underlying physical data generation processes.
Therefore, these two uncertainties are often modeled by probabilistic distributions over the model parameters and output space respectively. 

Earlier studies investigated uncertainty in arterial road travel time distributions \citep{ramezani2012travel}, or the multi-regime nature of traffic speed with Mixture-of-Expert formulations \citep{coric2011traffic}.
In addition, the idea of modeling uncertainty has also been studied in other time series forecasting tasks outside transportation domain, such as wind power prediction \citep{zheng2023stochastic} and medical data analysis \citep{tan2020explainable}.

With the rise of deep learning, Bayesian ideas such as Monte Carlo Dropout \citep{gal2016dropout} were combined with traffic forecasting models, for example, placing distributions \citep{wu2021bayesian} over the weights of a STGCN \citep{yu2018STGCN}.
Similar practices have also been proposed to develop a Bayesian Graph Convolution Network for spatial uncertainty modeling \citep{fu2021bstgcn}.
Later, both epistemic (weight) uncertainty and aleatoric uncertainty have been combined in one traffic prediction model \citep{qian2024uatgcn}, and then unified together with adaptive weight averaging \citep{qian2023towards}.
Frequentist approaches have emerged in parallel, for example, the quantile regression formulation has been applied to traffic prediction to adapt existing models into quantile predictors \citep{wu2023adaptive}.
To improve the generalization of uncertainty prediction, a spectral-normalization technique has been proposed to stabilize the training of Bayesian RNNs \citep{sengupta2024bayesian}.
Traffic forecasting and imputation have also been combined under one joint framework, where a copula-based multivariate distribution is learned to handle the joint distribution of multiple traffic variables \citep{an2025spatioprob}.

\modified{Although prior studies have approached the traffic forecasting problem through increasingly powerful spatio-temporal modeling and uncertainty-aware predictions, little emphasis has been placed on the multi-modality of traffic states, i.e., a road can be congested or uncongested.
Besides, probabilistic evaluation metrics often focus on data likelihood or the coverage of certain confidence intervals \citep{qian2023towards}, and lack further evaluation of the predictive distribution.
Therefore, in this work, we propose to address these gaps through a multi-modal probabilistic prediction module along with a comprehensive evaluation framework.
We adapt deterministic models with a GMM layer, allowing them to predict a multi-modal distribution to indicate different possible future modes.
For example, a GMM-based model can predict the probability of a road being congested or uncongested, along with the typical traffic speed of these two states. 
Our adaptation approach can be easily integrated into various model architectures while keeping data preparation, optimizer and hyperparameters unchanged.
Additionally, our evaluation framework evaluates the marginal predictive distribution through CRPS, and quantifies the quality of predictive intervals at multiple confidence levels through calibration and interval widths.
}

\section{Methodology}
\label{sec: methodology}
This section presents our straightforward approach to transform deterministic traffic forecasting models into effective probabilistic predictors using GMMs, and introduces the evaluation metrics to assess the quality of predicted distributions.

\subsection{Multi-Modal Probabilistic Traffic Forecasting}
\label{subsec: probabilistic traffic forecasting}
The goal of traffic forecasting is to predict the future traffic states at multiple future time steps over the transportation network, given the historical traffic states.
Since the transportation network naturally introduces spatial dependencies among locations, the traffic forecasting problem has often been modeled as a spatio-temporal prediction task.
In general, the structure of the neural networks can be described as follows:
\begin{equation}
\begin{aligned}
    \label{equ: deterministic method}
    \mathbf{Z} & = \text{ST\_Backbone} \left(\mathbf{X}_{t-T_h:t}, G\right)  \\
    \mathbf{X}_{t+1:t+T_f} & = \text{Det\_Pred}\left(\mathbf{Z} \right).
\end{aligned}
\end{equation}
ST\_Backbone($\cdot$) is a spatio-temporal modeling neural network (e.g., DCRNN \citep{li2018dcrnn} or GraphWaveNet \citep{wu2019graph}) that extracts hidden representations $\mathbf{Z} \in \mathbb{R}^{N \times D}$ using historical data and graph structure, where $D$ is the feature dimension.
In the input to the backbone, $\mathbf{X}_{t-T_h:t} \in \mathbb{R}^{N \times T_h}$ represents the traffic states for all $N$ locations across $T_h$ historical time steps up to the current time $t$.
Similarly, $\mathbf{X}_{t+1:t+T_f} \in \mathbb{R}^{N \times T_f}$ represents the future traffic states for all locations up to $T_f$ steps ahead of the current time $t$.
Following \citet{li2018dcrnn}, $T_f$ and $T_h$ are often set to an equal value $T$ in practice.
$\mathcal{G} = (\mathcal{V}, \mathcal{E})$ is the graph structure of the transportation network, whose nodes $\mathcal{V} = \{v_1, v_2, \ldots, v_N\}$ are the locations with collected traffic data, and edges $\mathcal{E}$ indicate the spatial dependencies among the locations.
The edges can often be equivalently transformed into an adjacency matrix $A \in \mathbb{R}^{N\times N}$, which can be binary for unweighted graphs and real-valued for weighted graphs.

Det\_Pred($\cdot$) is a deterministic layer that predicts the future traffic state values using the hidden features $\mathbf{Z}$ from the backbone.
Most forecasting methods predict $T$ future time steps in one forward pass \citep{lim2021time}, which is free from error accumulation in autoregressive multistep predictions.
As a result, the Det\_Pred layer can be described as:
\begin{equation}
    \mathbf{X}_{t+1:t+T_f} = \text{Linear}(\mathbf{Z}) = \mathbf{W} \mathbf{Z}^{\top} + \mathbf{b},
\end{equation}
where $\mathbf{W} \in \mathbb{R}^{T \times D}$ and $\mathbf{b} \in \mathbb{R}^{T}$ are the learnable weights and biases independently applied to each location.

As discussed in Section~\ref{sec: introduction}, a major limitation of the deterministic paradigm is that the model is always trained to predict a single most-likely value even when there are multiple possibilities.
Concretely, a road segment may develop into a congested state with low speed or a free-flow state with high traffic speed, which cannot be determined with current information.
Restricting the prediction space to a single value prevents the model from describing such complex uncertainty and hinders its ability to capture the stochastic nature of the data.
This limitation in descriptive power also creates challenges for prediction-based decision-making, as deterministic predictions may create an illusion of certainty and undermine system reliability.

\begin{figure}[ht]
    \centering
    \includegraphics[width=\linewidth]{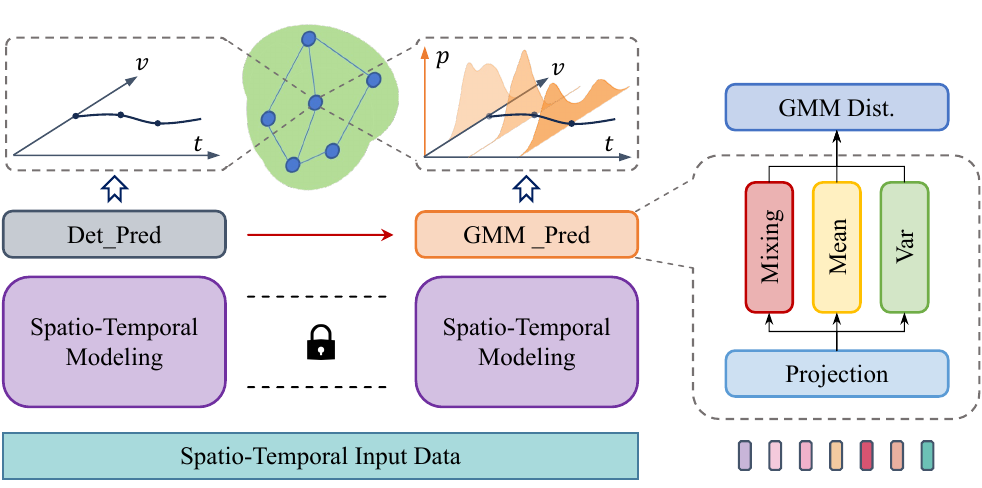}
    \caption{Transforming a deterministic model (left) to a probabilistic predictor (middle) with a GMM prediction layer (right).}
    \label{fig:gmm_adaptation}
\end{figure}

Therefore, this paper proposes to capture the diverse possibilities by modeling the traffic forecasting problem from a probabilistic perspective:
\begin{equation}
\begin{aligned}
    \mathbf{Z} & = \text{ST\_Backbone} \left(\mathbf{X}_{t-T_h:t}, G\right)  \\
    p(\mathbf{X}_{t+1:t+T_f})  & = \text{GMM\_Pred}\left(\mathbf{Z} \right),
\end{aligned}
\end{equation}
where $p(\mathbf{X}_{t+1:t+T_f})$ is the predicted probability distribution (instead of a point estimate) of future traffic states.
Despite the fundamental change in the prediction space, the adaptation can be achieved in a surprisingly straightforward manner.
As illustrated in Figure~\ref{fig:gmm_adaptation}, adapting a deterministic model requires no change in the spatio-temporal backbone.
The only modification is to replace the last output layer for deterministic prediction (Det\_Pred) with a GMM layer for multi-modal probabilistic predictions (GMM\_Pred)~\footnote{A Gaussian Mixture Model is called multi-modal because its probability density function can have multiple peaks, representing multiple modes.}.

\modified{The GMM\_Pred layer is a lightweight probabilistic enhancement module that can serve as a plug-in replacement for the deterministic prediction layer of existing traffic prediction models.}
Compared to the deterministic method, the GMM layer predicts a continuous multi-modal distribution, endowing the model with the ability to represent not only uncertainty around a mean value but also multiple distinct modes.
\modified{In this work, we do not consider the covariances across space and time}, and assume element-wise independent marginal distributions, which means the prediction at each location and time step is an independent one-dimensional Gaussian mixture distribution:
\begin{equation}
p(x) = \sum_{k=1}^{K} \pi_k \mathcal{N}(x ; \mu_k, \sigma_k^2),
\end{equation}
where $K$ is the number of components, $\mathcal{N}$ is the Normal density function, and $\pi_k$, $\mu_k$, and $\sigma_k^2$ are the predicted mixing coefficient, mean, and variance of the $k$-th component, respectively.
Notably, the mixing coefficients $\pi_k$ must sum to 1 to ensure the mixture is a valid probability distribution.

For the training loss, we use the GMM Negative Log-Likelihood (NLL) of future values (i.e., labels), whose element-wise form is:
\begin{equation}
\begin{aligned}
 -\log p(x) & = - \log \sum_{k=1}^{K} \pi_k \mathcal{N}(x ; \mu_k, \sigma_k^2) \\
            & = - \log \sum_{k=1}^{K} \exp \left( \log  \left(\pi_k \mathcal{N}(x ; \mu_k, \sigma_k^2)\right) \right) \\
            & = - \log \sum_{k=1}^{K} \exp \left( \log \pi_k - \frac{(x - \mu_k)^2}{2\sigma_k^2}  -  \frac{1}{2} \log(2\pi \sigma_k^2) \right), \\
\end{aligned}
\label{equ: gmm nll loss}
\end{equation}
where $x$ is the value of an observed traffic variable at a specific location and time step.
The element-wise loss is then averaged over all locations and time steps to obtain the final training loss, without any auxiliary terms.

\modified{In fact, the idea of the GMM layer is closely related to the Mixture Density Network (MDN) \citep{bishop1994mixture}, which trains neural networks to predict parameterized mixture distributions.
We adapt this idea to the traffic forecasting task, and demonstrate through extensive experiments across various spatio-temporal models that such a probabilistic upgrade is straightforward, efficient and practically applicable.}

\subsection{Details of the GMM Layer}
\label{subsec: gmm layer}
As illustrated in Figure~\ref{fig:gmm_adaptation}, the GMM layer starts with a projection layer to transform location-specific features, implemented as a Linear layer:
\begin{equation}
    \mathbf{Z}' = \text{Linear} (\mathbf{Z}),
\end{equation}
where a linear transform is independently applied to each location, and $\mathbf{Z}' \in \mathbb{R}^{N \times D'}$ is the transformed feature matrix with hidden dimension $D'$.
Then, $\mathbf{Z}'$ is fed into three separate branches for the mixing coefficients, means, and variances to build the mixture distribution.
\begin{equation}
    \begin{aligned}
        \boldsymbol{\pi} & = \text{Softmax}\left(\text{Reshape}\left(\text{Linear} (\mathbf{Z}')\right)\right), \\
        \boldsymbol{\mu} & = \text{Reshape}\left(\text{Linear} (\mathbf{Z}')\right) \cdot s + \mathbf{r}, \\
        \log \left( \boldsymbol{\sigma}^2 \right) & = \text{Reshape}\left(\text{Linear} (\mathbf{Z}')\right). \\
    \end{aligned}
\end{equation}
The Linear($\cdot$) layer in each branch predicts the corresponding parameters for all $K$ components at all $T$ future time steps simultaneously, resulting in output shapes of $\mathbb{R}^{N \times (T \times K)}$ for each branch.
Then, the Reshape($\cdot$) operation reshapes the output into $\mathbb{R}^{N \times T \times K}$.

To ensure the mixture is a valid probability distribution, the mixing coefficients $\boldsymbol{\pi} \in \mathbb{R}^{N \times T \times K}$ are normalized over the $K$ components using the Softmax function, whose element-wise formula (at a specific location and time step) is:
\begin{equation}
    \pi_k = \frac{\exp(z_k)}{\sum_{j=1}^{K} \exp(z_j)},
\end{equation}
where $z_k$ is the output of the linear layer in the mixing-coefficient branch for the $k$-th Gaussian component.
Following \citet{kendall2017uncertainties}, the GMM layer predicts log-variance instead of the variance itself for better numerical stability, i.e., $\log \left(\boldsymbol{\sigma}^2\right) \in \mathbb{R}^{N \times T \times K}$.
The mean values $\boldsymbol{\mu} \in \mathbb{R}^{N \times T \times K}$ are re-parameterized by the predicted relative offsets together with pre-defined (not learnable) reference values $\mathbf{r} \in \mathbb{R}^{K}$ and scale $s \in \mathbb{R}$.
The $\cdot$ between the scale $s$ and transformed features is a simple multiplication that applies the same scale factor to all locations and time steps.
This formulation can inject prior knowledge about the data distribution and stabilize training.

In our implementation, both the inputs and labels in Equation~\ref{equ: gmm nll loss} are normalized by a Z-score transformation to have zero mean and unit variance, and the training loss is computed in normalized space.
The final form of the GMM NLL loss utilizes the log-sum-exp approach to avoid numerical instability.
In addition, the GMM layer is initialized to predict a weakly informative prior \citep{gelman1995bayesian}, where all mixing coefficients are equal, the means are the reference values (zero offsets), and the variances are set to one.
This initialization involves both neural network parameters and the reference values and scale in the mean-value formulation.
Specifically, the linear layer of the mixing branch is initialized with zero weights and a bias of $1/K$, while the mean and variance branches are both initialized with zero weights and zero bias.

\begin{figure}[ht]
    \centering
    \includegraphics[width=0.8\linewidth]{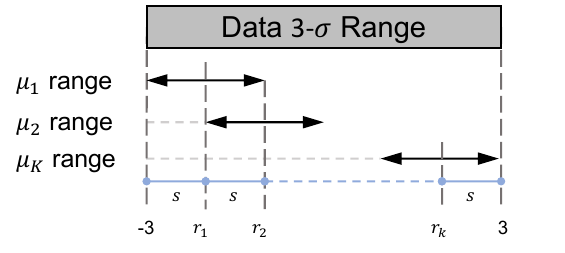}
    \caption{Mean value formulation and data range}
    \label{fig:gmm_init_params}
\end{figure}

\modified{
Figure~\ref{fig:gmm_init_params} illustrates the intuition for the mean value formulation, whose purpose is to allow the predicted component mean to have sufficient coverage over the data range.
Since the data are normalized through a Z-score transform, the 3-$\sigma$ range is approximately $[-3, 3]$.
The reference values $\mathbf{r} = [r_1, ..., r_K]$ are evenly placed in this range, with the spacing being $s$.
With this formulation, we have
\begin{equation}
    -3 + (K+1)s= 3 \rightarrow (K+1)s = 6.
\end{equation}
As a result, a unit-range output for the first Gaussian component will be mapped to $[r_1 - s, r_1 + s]$ after scaling and offset, and a ground-truth value can be approached by a predicted component mean (from the closest reference value), without requiring the underlying linear layer to generate a large output (more details in \ref{app: reference values and scales}).}
We use a default of $K=5$ mixture components in the GMM layer to allow more flexibility in modeling the data distribution, and therefore $s=1$.
Then, the corresponding reference values are set to be $\mathbf{r} = [-2, -1, 0, 1, 2]$.

\subsection{Derived Predictions}
\label{subsec: derived predictions}
From the predicted distribution, we extract point estimates and interval-based predictions to demonstrate the versatility of our multi-modal probabilistic predictions across various application scenarios.
The Bayesian average of the predicted distribution can serve as a point estimate, which is the weighted sum of the component means:
\begin{equation}
    \hat{x} = \sum_{k=1}^{K} \pi_k \mu_k.
    \label{equ: gmm bayesian average point estimation}
\end{equation}

\begin{figure}[ht]
    \centering
    \includegraphics[width=\linewidth]{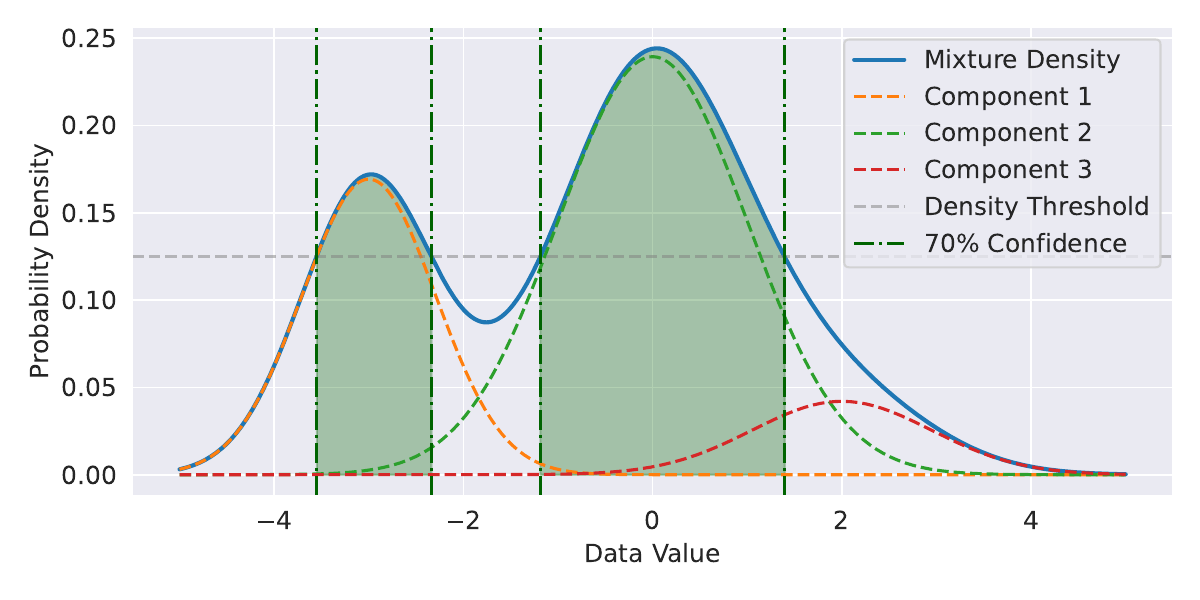}
    \caption{Example HDPS for GMM. The 70\% HDPS consists of two sub-intervals under the shaded area.}
    \label{fig:gmm_density_interval}
\end{figure}

\modified{We refer to our interval-based prediction as High-Density Prediction Sets (HDPS), i.e., a set of sub-intervals at a certain confidence level while prioritizing high-density regions.}
The procedure is described in Algorithm~\ref{alg: hdps derivation}.
\modified{Since this work targets traffic speed prediction, the range of interest is from 0 to the speed limit of each dataset, and the numerical density is evaluated with 2000 points evenly spaced in this range.}
Figure~\ref{fig:gmm_density_interval} illustrates an example HDPS using a three-component GMM.
For intuition, one can imagine a density threshold, i.e., the horizontal line in Figure~\ref{fig:gmm_density_interval}, intersecting the density curve.
This threshold line may have multiple intersections with the density curve, and each resulting segment corresponds to a sub-interval at the given confidence level.
When the Gaussian components are well-separated and have small variances, the density curve for a GMM with $K$ components can have $K$ peaks.
As a result, the threshold line can have up to $K$ line segments inside the density curve, yielding up to $K$ sub-intervals for a given confidence level.

\begin{algorithm}[ht]
\caption{Derivation of HDPS}
\label{alg: hdps derivation}
\begin{algorithmic}[1]
\STATE \textbf{Input:} GMM $\{\pi_k, \mu_k, \sigma_k\}_{k=1}^K$, confidence level $c$
\STATE \textbf{Output:} HDPS $\{l^k, u^k\}_{k}$
\STATE Calculate numerical density $p(x)$ at multiple points $x$. \\
$x$ has step size $\Delta x$ and covers the range of interest
\STATE Sort $p(x)$ by descending order of probability density
\STATE Compute a cumulative sum for the sorted density \\
    $F(x) = \operatorname{cumsum}(p(x)*\Delta x)$
\STATE Normalize: $F(x) \leftarrow F(x) / \operatorname{max}(F(x))$
\STATE Search the insertion index $i$ for $c$ in $F(x)$
\STATE All elements to the left of $i$ are selected (high-density)
\STATE Backtrack to the original $p(x)$ to find selected $x$ values
\STATE Find $\{l^k\}$ and $\{u^k\}$ as follows: \\
$\{l^k\}$: selected $x$'s where $x-\Delta x$ is not selected \\
$\{u^k\}$: selected $x$'s where $x+\Delta x$ is not selected \\
\STATE Sort $\{l^k\}$ and $\{u^k\}$ and pair by index
\STATE \textbf{Return} $\{l^k, u^k\}_{k}$
\end{algorithmic}
\end{algorithm}

\subsection{Evaluating Probabilistic Predictions}
\label{subsec: evaluating probabilistic predictions}
\modified{In this section, we propose a suite of metrics for evaluating probabilistic traffic predictors.}
Since a probabilistic prediction contains much more information than a deterministic counterpart, ideal scoring rules should evaluate the entire distribution rather than derived properties.
For this purpose, we follow \citet{tang2021probabilistic} and use the Continuous Ranked Probability Score (CRPS) \citep{gneiting2007strictly} for its joint consideration of calibration, resolution, and uncertainty \citep{candille2005evaluation}.
By definition, the CRPS score evaluates the difference between a predicted Cumulative Distribution Function (CDF) $F$ and an observed value $y$:
\begin{equation}
    \operatorname{CRPS}(F, y)=\int_{-\infty}^{\infty}\left(F(x)-H(x - y)\right)^2 dx,
    \label{equ: crps}
\end{equation}
where $H$ is the Heaviside step function, i.e., a step function jumping from 0 to 1 at the label value $y$.
We interpret a deterministic prediction as a Dirac Delta function.
Therefore, $F(x)$ reduces to a step CDF at the predicted value, and the CRPS score becomes the absolute error between the prediction and the label.
In addition to CRPS, we introduce two metrics to specifically evaluate the interval-based predictions, i.e., the mean Average Width (mAW) and the mean Confidence Calibration Error (mCCE).
\modified{Below, we describe a general notation to integrate HDPS into the concept of confidence interval, where a predicted confidence interval consists of multiple sub-intervals, and a single continuous interval can be regarded as a special case.}

A properly predicted confidence interval should be informative.
Therefore, we measure the width of the predicted interval (possibly with multiple sub-intervals indexed by $i$) at a confidence level, and average it over all locations and time steps to obtain an Average Width ($\operatorname{AW}$).
\begin{equation}
    \operatorname{AW} = \frac{1}{|T||V|} \sum_{t,v} \left( \sum_{i} |u_{tv}^i - l_{tv}^i| \right),
\end{equation}
where $|T|$ and $|V|$ are the number of prediction time steps and locations, and $u_{tv}^i$ and $l_{tv}^i$ are the upper and lower bounds of the $i$-th predicted sub-interval at time step $t$ and location $v$.
Then, AW is averaged over all confidence levels of interest to obtain mAW.
\begin{equation}
    \operatorname{mAW} = \frac{1}{|C|} \sum_{c} \operatorname{AW}_c,
    \label{equ: maw}
\end{equation}
where $|C|$ is the number of confidence levels, and $\operatorname{AW}_c$ is the average width at confidence level $c$.
\modified{In this work, we consider 10 confidence levels from 0.5 to 0.95 with a step size of 0.05}.

While the mAW measures how informative the confidence intervals are, the predicted intervals should also be well-calibrated, e.g., a 90\% confidence interval should contain the true label 90\% of the time.
When the confidence and coverage are misaligned (either over-confident or under-confident), the predicted intervals can mislead decision processes relying on those predictions.
In fact, a confidence interval can be regarded as a binary classification, where the model predicts that the label value is within the interval with a certain confidence level.
Therefore, we draw inspiration from \citet{guo2017calibration} and compute the Confidence Calibration Error (CCE) at a confidence level as the absolute difference between the expected confidence level and the empirical coverage:
\begin{equation}
    \label{equ: cce}
    \operatorname{CCE} = \left| \frac{1}{|T||V|} \sum_{t,v} \left( \sum_{i} \mathbf{1}(y_{tv} \in [l_{tv}^i, u_{tv}^i]) \right)  - c \right|,
\end{equation}
where $\mathbf{1}(\cdot)$ is the indicator function that returns 1 if the condition ($y_{tv} \in [l_{tv}^i, u_{tv}^i]$) is satisfied, and 0 otherwise.
In the case of multiple sub-intervals, the true label is considered to be within the interval if it falls into any of the sub-intervals.
Similar to the metric mAW, we also average the CCE over the same set of confidence levels to obtain the mCCE:
\begin{equation}
    \label{equ: mcce}
    \operatorname{mCCE} = \frac{1}{|C|} \sum_{c} \operatorname{CCE}_c.
\end{equation}

\section{Experiments}
\label{sec: experiments}

This section first presents the experimental setup, then introduces the evaluation results using probabilistic metrics and traditional deterministic metrics, and finally discusses the impact of data quality on uncertainty through an additional study.

\subsection{Datasets and Experiment Setup}
\label{subsec: datasets and experiment setup}

The experiments and evaluations are conducted on two widely recognized datasets, METR-LA and PEMS-Bay \citep{li2018dcrnn}, and a novel dataset, SimBarcaSpd, derived from the SimBarca dataset with dense urban traffic from 1570 road segments in a 15 $km^2$ area of central Barcelona \citep{xiong2025multi}.
The dense connections create more complex spatial dependencies, thus making urban traffic forecasting much more challenging than traditional datasets with sparser road networks.
Unlike METR-LA and PEMS-Bay, which are collected from highway loop detectors, SimBarcaSpd is derived from a microscopic traffic simulator using the dense road network of central Barcelona.
All three datasets provide traffic speed data, whose maximum values differ due to regional speed limits (the minimum is naturally 0).
At the same time, error metrics are computed in the same units as those maximum values.
These specifications are summarized in Table~\ref{tab:dataset_specs}.
\begin{table}[ht]
\centering
\setlength{\tabcolsep}{1mm}
\caption{Dataset specifications}
\label{tab:dataset_specs}
\resizebox{\columnwidth}{!}{%
\begin{tabular}{ccccccc}
    \toprule
    Dataset & In-Step & Out-Step & Step Size & \#Nodes & Max & Area\\
    \midrule
    METR-LA & 12 & 12 & 5 min & 207 & 70 mph & 477 $km^2$\\
    PEMS-Bay & 12 &  12 & 5 min & 325 & 85 mph & 325 $km^2$\\
    SimBarcaSpd & 10 & 10 & 3 min & 1570 & 14 m/s & 15 $km^2$\\
    \bottomrule
\end{tabular}
}
\end{table}

\begin{figure}[ht]
    \centering
    \includegraphics[width=0.8\linewidth]{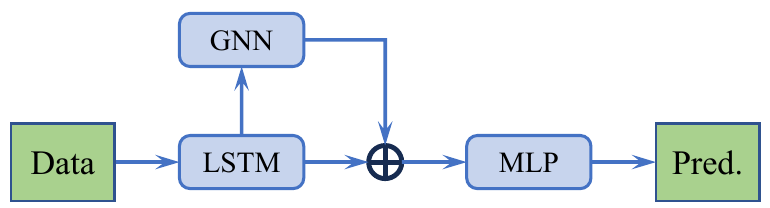}
    \caption{Structure of the LGC model}
    \label{fig:lgc_model}
\end{figure}

To show that our approach can be applied to a wide range of traffic forecasting models, we conduct experiments on several representative models, including GWNet \citep{wu2019graph}, MTGNN \citep{wu2020connecting}, STID \citep{shao2022spatial}, STAEformer \citep{liu2023spatio}, and a simple baseline model LSTM-GCN (denoted as LGC), as shown in Figure~\ref{fig:lgc_model}.
The LGC model first encodes the temporal dimension using a three-layer LSTM \citep{hochreiter1997long} and then processes spatial dependencies using a three-layer Graph Neural Network implemented with GCNConv \citep{kipf2016semi}.
Then, the spatial and temporal features are concatenated and passed to a Multi-Layer Perceptron (MLP).

For each of these models, three variants are considered: a deterministic variant (Det), a Normal variant (Norm) predicting a Gaussian distribution, and a GMM variant with the adaptation method in Figure~\ref{fig:gmm_adaptation}.
In fact, the Norm variant is a special case of the GMM variant with a single component.

For a fair comparison, we use a consistent training and evaluation pipeline across all models and variants.
The models are trained for 50 epochs with a batch size of 32, except STAEformer, which is trained for 20 epochs with a batch size of 8 to fit memory constraints and prevent heavy overfitting.
The AdamW optimizer \citep{loshchilov2018decoupled} is used with a learning rate of 0.0005, weight decay of 0.0001 and beta parameters of (0.9, 0.999).
The learning rate gradually increases from 0 to 0.0005 in the first 2 epochs, and decays to 10\% and 1\% of the chosen learning rate at 75\% and 85\% progress through the total training epochs.
\modified{In evaluation, the numerical probability density in Algorithm~\ref{alg: hdps derivation} is computed at 2000 evenly spaced points from 0 to the maximum speed of each dataset in Table~\ref{tab:dataset_specs}.}
Unless otherwise noted, the accuracy experiments are conducted on a single Nvidia L40S GPU with 48GB memory, and the implementation is based on PyTorch.

\subsection{Overall Evaluation Results}
\label{subsec: overall evaluation results}

Table~\ref{tab:probabilistic_performance_metrics} presents the probabilistic performance metrics, i.e., CRPS, mAW and mCCE, for all three datasets. 
The metrics are averaged over all prediction horizons to provide a joint evaluation of model performance at both shorter and longer horizons.
The rows are grouped by the architecture of base models, with individual variants (Det, Norm and GMM) listed separately.
As discussed in the methodology, the CRPS for the Det variant is the same as the Mean Absolute Error (MAE) in the deterministic prediction task.
Moreover, mAW and mCCE for the Det variants are not applicable, as these variants do not provide confidence intervals.

On the primary metric, CRPS, a consistent hierarchy is observed across all datasets and all base model architectures, where the GMM variant achieves the best performance, followed by the Norm variant, and the Det variant has the worst performance.
The difference between the tiers is significant, which shows the advantage of probabilistic predictions over deterministic ones, and the benefit of multi-modal predictions.
STAEformer (briefly noted as STAE) achieves the best CRPS scores on all three datasets, due to its advanced spatio-temporal modeling architecture.
Another notable observation is that the performance gap between different variants is more pronounced than the gap induced by base model architectures.
For example, the variant STAEformer-GMM achieves a CRPS of 2.23 on METR-LA, which is much better than its Norm variant (2.37) and Det variant (3.04).
Still, the GMM variant of simpler models, such as LGC and GWNet, can also achieve similar CRPS scores at 2.25 and 2.26, respectively.
This observation shows that the probabilistic modeling approach has a stronger impact on prediction performance.

The experiments clearly demonstrate the effectiveness and generality of our proposed GMM adaptation method across different backbone architectures.
Using the deterministic variant as a baseline (100\%), we observe in Table~\ref{tab:crps_improvement_percent} a consistent and significant average relative CRPS improvement of 27.7\% by introducing the GMM layer.

The confidence intervals are evaluated using both mAW and mCCE, and ideal predictions should have low values for both metrics simultaneously.
\modified{On nearly all datasets and base architectures, the GMM variants achieve equal or better mAW and mCCE scores than their Norm variants, which means the predicted intervals are more informative and better calibrated at the same time.
The only exception is STAEformer on the PEMS-Bay dataset, where the GMM variant predicts more informative intervals (smaller mAW) and the Norm variant is better calibrated.
These results suggest that the proposed GMM adaptation method can be effective and suitable for various scenarios.
}

\begin{table}[ht]
\centering
\setlength{\tabcolsep}{1mm}
\caption{Probabilistic performance metrics. Lower values are preferred for all, best result for a backbone in \textbf{bold}.
\modified{Due to numerical density estimation for GMM, a small difference in the metrics may not have enough significance, e.g., $\sim$0.002 CRPS, $\sim$0.005 mCCE or $\sim$0.05 mAW, please check \ref{app: numerical accuracy of density estimation} for details}}
\label{tab:probabilistic_performance_metrics}
\setlength{\tabcolsep}{1mm}
\resizebox{\columnwidth}{!}{%
\begin{tabular}{cc|ccc|ccc|ccc}
    \toprule
    \multicolumn{2}{c|}{\multirow{2}{*}{Method}}
    & \multicolumn{3}{c|}{METR-LA}
    & \multicolumn{3}{c|}{PEMS-Bay}
    & \multicolumn{3}{c}{SimBarcaSpd} \\
    \cmidrule(lr){3-5} \cmidrule(lr){6-8} \cmidrule(lr){9-11}
    \multicolumn{2}{c|}{}
    & \rotatebox{90}{CRPS} & \rotatebox{90}{mAW}  & \rotatebox{90}{mCCE}
    & \rotatebox{90}{CRPS} & \rotatebox{90}{mAW}  & \rotatebox{90}{mCCE}
    & \rotatebox{90}{CRPS} & \rotatebox{90}{mAW}  & \rotatebox{90}{mCCE}        \\
    \midrule
    
    \multirow{3}{*}{\rotatebox{90}{LGC}}
    & Det  & 3.11 & -     & -    & 1.65 & -    & -    & 1.05 & -    & -    \\
    & Norm & 2.49 & 9.61  & 0.06 & 1.29 & 5.12 & \textbf{0.04} & 0.81 & 3.08 & 0.03 \\
    & GMM  & \textbf{2.25} & \textbf{7.14}  & \textbf{0.01} & \textbf{1.22} & \textbf{3.89} & \textbf{0.04} & \textbf{0.75} & \textbf{2.49} & \textbf{0.02} \\
    \midrule
    
    \multirow{3}{*}{\rotatebox{90}{STID}}
    & Det  & 3.31 & -     & -    & 1.67 & -    & -    & 1.06 & -    & -    \\
    & Norm & 2.68 & 10.67 & 0.09 & 1.31 & 5.94 & 0.08 & 0.82 & 3.28 & 0.05 \\
    & GMM  & \textbf{2.38} & \textbf{7.71}  & \textbf{0.00} & \textbf{1.22} & \textbf{4.36} & \textbf{0.00} & \textbf{0.76} & \textbf{2.58} & \textbf{0.01} \\
    \midrule
    
    \multirow{3}{*}{\rotatebox{90}{GWNet}}
    & Det  & 3.13 & -     & -    & 1.58 & -    & -    & 1.04 & -    & -    \\
    & Norm & 2.52 & 9.77  & 0.07 & 1.23 & 5.19 & 0.05 & 0.81 & 3.19 & 0.03 \\
    & GMM  & \textbf{2.26} & \textbf{7.19}  & \textbf{0.01} & \textbf{1.14} & \textbf{3.92} & \textbf{0.02} & \textbf{0.76} & \textbf{2.62} & \textbf{0.01} \\
    \midrule
    
    \multirow{3}{*}{\rotatebox{90}{MTGNN}}
    & Det  & 3.12 & -     & -    & 1.57 & -    & -    & 1.00 & -    & -    \\
    & Norm & 2.54 & 10.03 & 0.07 & 1.22 & 5.33 & 0.06 & 0.78 & 3.08 & 0.04 \\
    & GMM  & \textbf{2.27} & \textbf{7.27}  & \textbf{0.01} & \textbf{1.15} & \textbf{3.95} & \textbf{0.02} & \textbf{0.72} & \textbf{2.45} & \textbf{0.01} \\
    \midrule
    
    \multirow{3}{*}{\rotatebox{90}{STAE}}
    & Det  & 3.04 & -     & -    & 1.56 & -    & -    & 0.98 & -    & -    \\
    & Norm & 2.37 & 8.34  & \textbf{0.02} & 1.17 & 4.33 & \textbf{0.01} & 0.74 & 2.93 & 0.04 \\
    & GMM  & \textbf{2.23} & \textbf{6.67}  & \textbf{0.02} & \textbf{1.13} & \textbf{3.52} & 0.06 & \textbf{0.69} & \textbf{2.37} & \textbf{0.00} \\
    \bottomrule
\end{tabular}
}
\end{table}

\begin{table}[ht]
\centering
\caption{Relative CRPS improvement (\%) of GMM over deterministic variants}
\label{tab:crps_improvement_percent}
\setlength{\tabcolsep}{1mm}
\resizebox{\columnwidth}{!}{%
\begin{tabular}{l|ccccc|c}
\toprule
Dataset & LGC & STID & GWNet & MTGNN & STAE & Avg. \\
\midrule
METR-LA     & 27.74 & 28.31 & 28.03 & 27.33 & 28.16 & 27.91 \\
PEMS-Bay    & 26.95 & 27.38 & 27.85 & 26.75 & 28.30 & 27.45 \\
SimBarcaSpd & 28.57 & 27.62 & 26.92 & 27.27 & 28.87 & 27.85 \\
\midrule
Average     & 27.75 & 27.77 & 27.60 & 27.12 & 28.44 & 27.74 \\
\bottomrule
\end{tabular}
}
\end{table}

Table~\ref{tab:deterministic_performance_metrics} presents the deterministic performance metrics, as is common practice in the literature \citep{li2018dcrnn}.
The metrics include Mean Absolute Error (MAE), Mean Absolute Percentage Error (MAPE) and Root Mean Square Error (RMSE).
When comparing across base models, STAEformer achieves the best deterministic performance on all three datasets, which is consistent with the literature \citep{liu2023spatio}.
In general, the GMM variants achieve better RMSE, and the natively deterministic models have better MAE and MAPE.
This can be attributed to the fact that the deterministic models are trained with MAE as the loss function, and the GMM NLL loss in Equation~\ref{equ: gmm nll loss} has an L2 form closer to the Mean Square Error (the square of RMSE).
Since our experiments do not change the hyperparameters or backbone model architectures, it is reasonable that the GMM variants usually perform better on RMSE while the deterministic ones are better on MAE and MAPE.

\begin{table}[ht]
\centering
\caption{Deterministic performance metrics. Lower values are preferred for all, best result for a backbone in \textbf{bold}.}
\label{tab:deterministic_performance_metrics}
\setlength{\tabcolsep}{1mm}
\resizebox{\columnwidth}{!}{%
\begin{tabular}{cc|ccc|ccc|ccc}
\toprule
\multicolumn{2}{c|}{\multirow{2}{*}{Method}}
    & \multicolumn{3}{c|}{METR-LA}
    & \multicolumn{3}{c|}{PEMS-Bay}
    & \multicolumn{3}{c}{SimBarcaSpd}  \\
\cmidrule(lr){3-5} \cmidrule(lr){6-8} \cmidrule(lr){9-11}
\multicolumn{2}{c|}{}
    & \rotatebox{90}{MAE}  & \rotatebox{90}{MAPE} & \rotatebox{90}{RMSE}  
    & \rotatebox{90}{MAE}  & \rotatebox{90}{MAPE} & \rotatebox{90}{RMSE} 
    & \rotatebox{90}{MAE}  & \rotatebox{90}{MAPE} & \rotatebox{90}{RMSE}         \\
\midrule

\multirow{3}{*}{\rotatebox{90}{LGC}}
    & Det  & \textbf{3.11} & \textbf{8.69}  & 6.20 & \textbf{1.65} & \textbf{3.92} & \textbf{3.60} & \textbf{1.05} & \textbf{24.06} & 1.76 \\
    & Norm & 3.35 & 9.42  & 6.13 & 1.78 & 4.33 & 3.69 & 1.12 & 25.91 & 1.71 \\
    & GMM  & 3.24 & 9.08  & \textbf{6.01} & 1.73 & 4.30 & 3.68 & 1.10 & 25.93 & \textbf{1.70} \\
\midrule

\multirow{3}{*}{\rotatebox{90}{STID}}
    & Det  & \textbf{3.31} & \textbf{9.59}  & 6.68 & \textbf{1.67} & \textbf{3.81} & 3.70 & \textbf{1.06} & \textbf{24.10} & 1.77 \\
    & Norm & 3.55 & 10.15 & 6.46 & 1.77 & 4.04 & 3.69 & 1.12 & 25.70 & 1.73 \\
    & GMM  & 3.46 & 9.88  & \textbf{6.37} & 1.73 & 3.97 & \textbf{3.64} & 1.11 & 25.59 & \textbf{1.72} \\
\midrule

\multirow{3}{*}{\rotatebox{90}{GWNet}}
    & Det  & \textbf{3.13} & \textbf{8.70}  & 6.27 & \textbf{1.58} & \textbf{3.59} & 3.52 & \textbf{1.04} & \textbf{24.00} & 1.74 \\
    & Norm & 3.38 & 9.42  & 6.17 & 1.67 & 3.79 & 3.50 & 1.13 & 26.07 & \textbf{1.71} \\
    & GMM  & 3.28 & 9.12  & \textbf{6.08} & 1.61 & 3.65 & \textbf{3.44} & 1.11 & 25.81 & \textbf{1.71} \\
\midrule

\multirow{3}{*}{\rotatebox{90}{MTGNN}}
    & Det  & \textbf{3.12} & \textbf{8.64}  & 6.37 & \textbf{1.57} & \textbf{3.53} & 3.52 & \textbf{1.00} & \textbf{22.91} & 1.69 \\
    & Norm & 3.40 & 9.36  & 6.18 & 1.66 & 3.77 & 3.46 & 1.08 & 25.14 & \textbf{1.65} \\
    & GMM  & 3.28 & 9.08  & \textbf{6.10} & 1.62 & 3.69 & \textbf{3.45} & 1.06 & 24.63 & \textbf{1.65} \\
\midrule

\multirow{3}{*}{\rotatebox{90}{STAE}}
    & Det  & \textbf{3.04} & \textbf{8.36}  & 6.13 & \textbf{1.56} & \textbf{3.45} & 3.51 & \textbf{0.98} & \textbf{22.81} & 1.67 \\
    & Norm & 3.19 & 8.74  & \textbf{5.97} & 1.61 & 3.64 & 3.42 & 1.02 & 23.91 & \textbf{1.59} \\
    & GMM  & 3.12 & 8.43  & \textbf{5.97} & 1.58 & 3.50 & \textbf{3.37} & 1.02 & 23.66 & \textbf{1.59} \\
\bottomrule

\end{tabular}
}
\end{table}

\subsection{Performance Analysis by Prediction Horizons}
\label{subsec: performance Analysis by prediction horizons}

\begin{figure*}[ht]
    \centering
    \subfloat[METR-LA]{\includegraphics[width=0.32\textwidth]{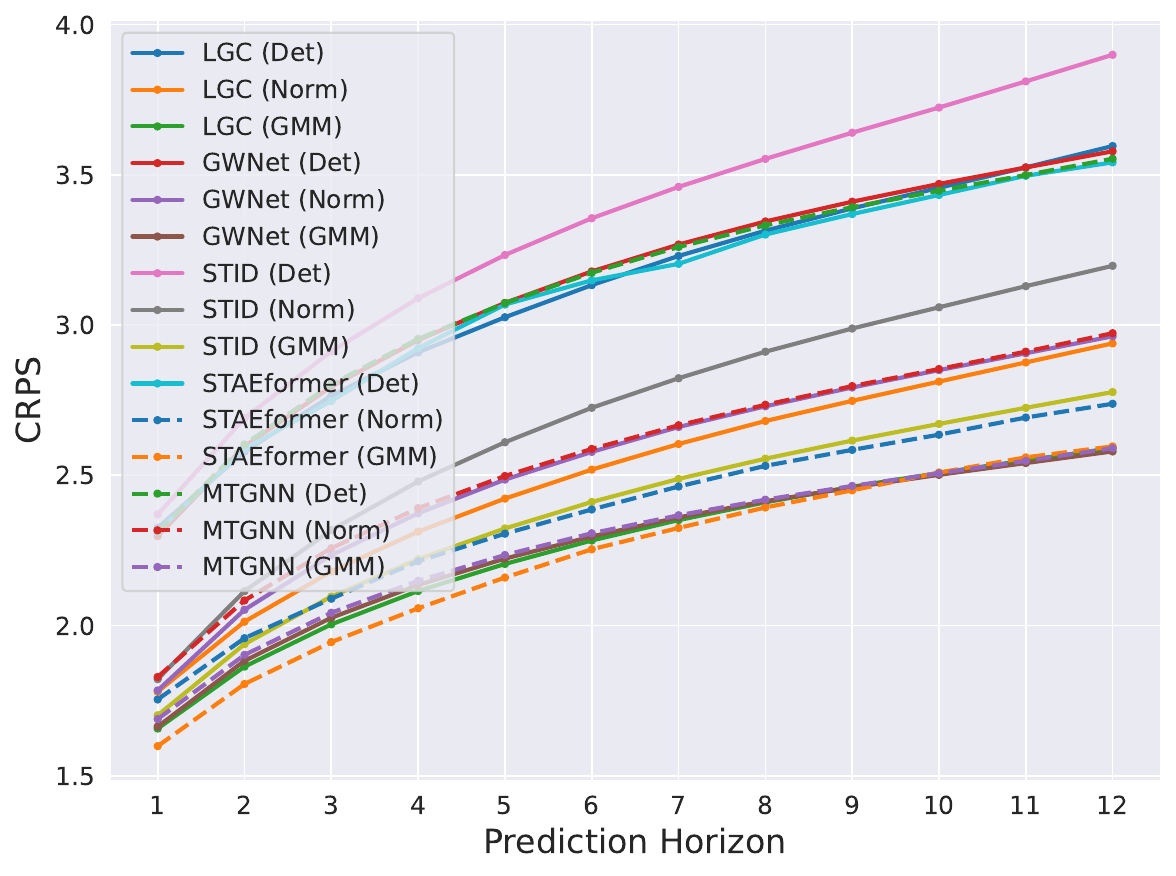}}
    \hfill
    \subfloat[PEMS-Bay]{\includegraphics[width=0.32\textwidth]{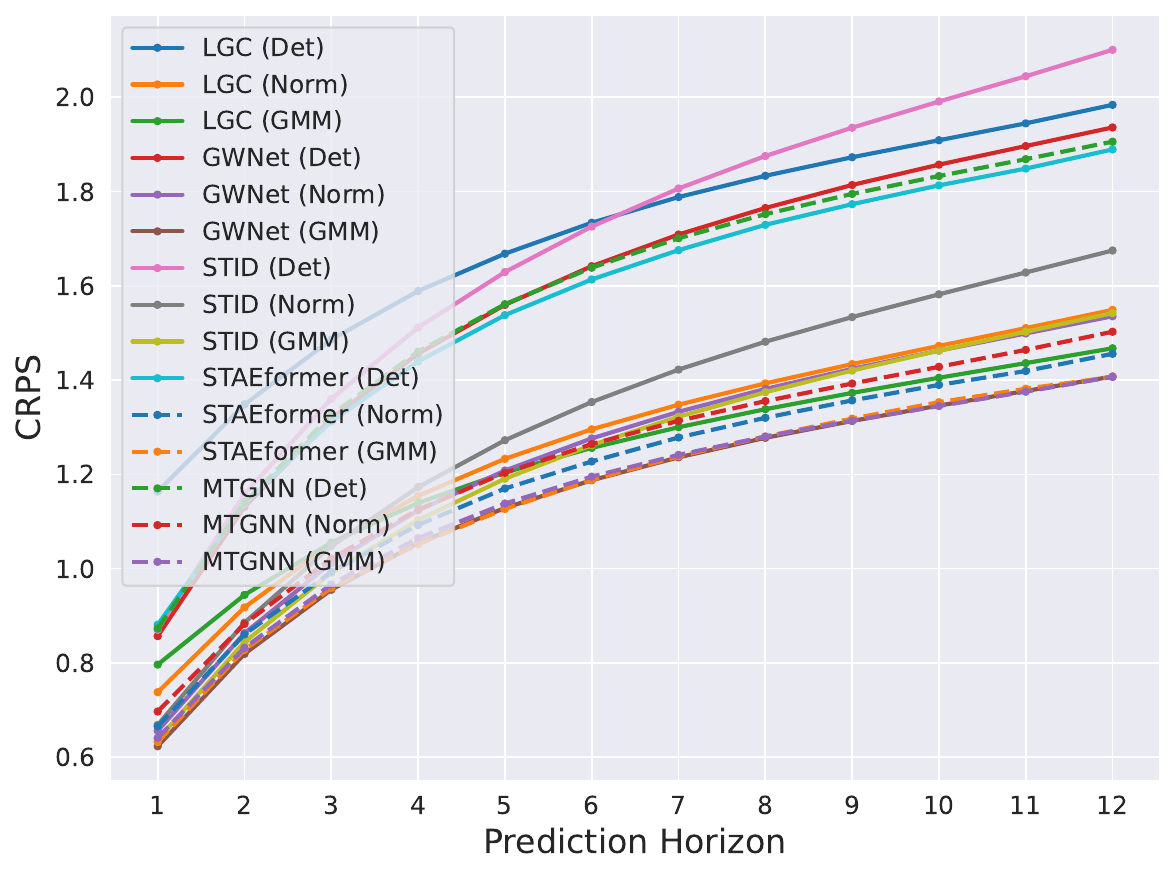}}
    \hfill
    \subfloat[SimBarcaSpd]{\includegraphics[width=0.32\textwidth]{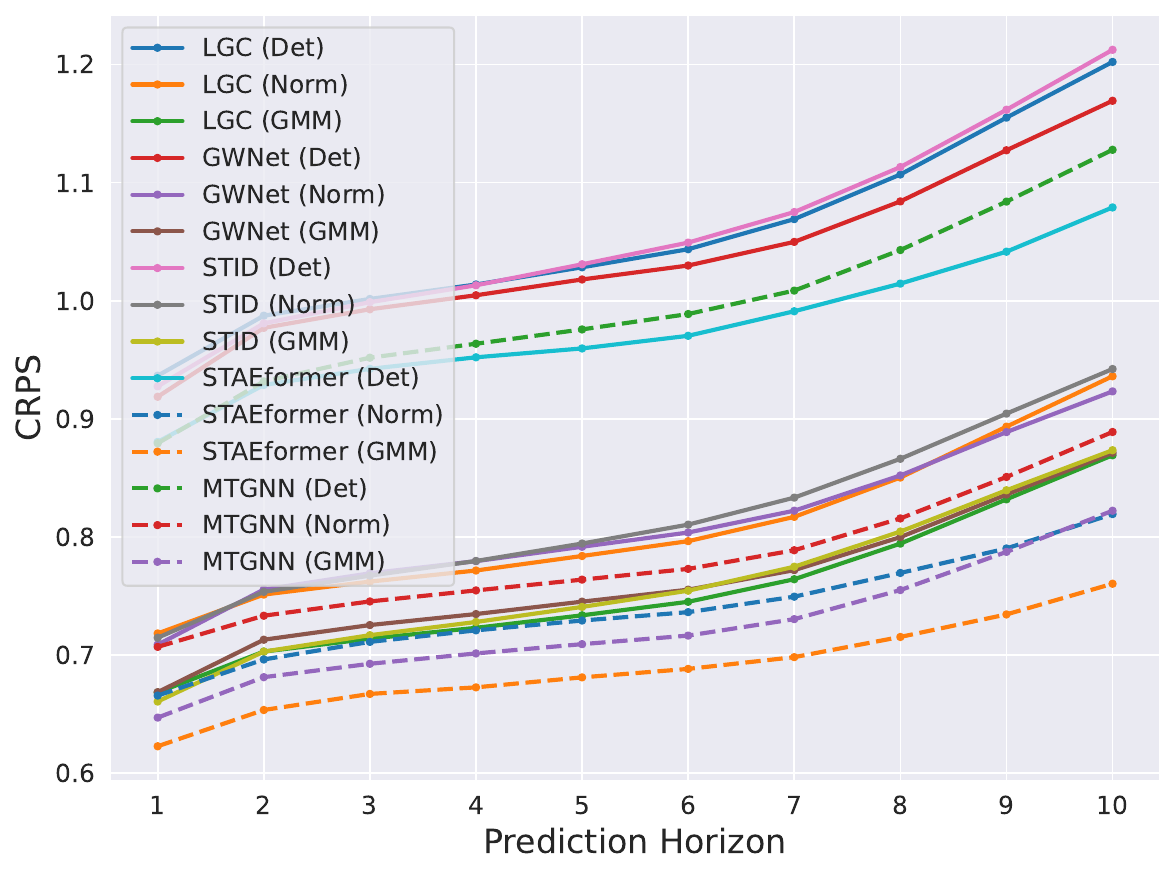}}
    \caption{CRPS score of all methods at all prediction horizons on three datasets. \modified{The figures in this section and Section~\ref{subsec: prediction examples} are provided at high resolution, please zoom in for details}}
    \label{fig:crps_scores_by_prediction_horizon}
\end{figure*}

Figure~\ref{fig:crps_scores_by_prediction_horizon} shows the CRPS scores of all methods at all prediction horizons, providing a more detailed view of the CRPS results in Table~\ref{tab:probabilistic_performance_metrics}.
A clear trend is that CRPS increases with prediction horizon for all methods, as predictions become more challenging.
From the figures, a clear gap can be identified between the deterministic and probabilistic variants, which again demonstrates the drawbacks of false confidence and the necessity of probabilistic modeling in traffic forecasting.
The performance hierarchy is mostly consistent across all prediction horizons, where a stronger backbone model (e.g. STAEformer) consistently outperforms a weaker one (e.g. STID), and the GMM variant consistently outperforms the Normal variant.

\begin{figure*}[ht]
    \centering
    \subfloat[METR-LA]{\includegraphics[width=0.32\textwidth]{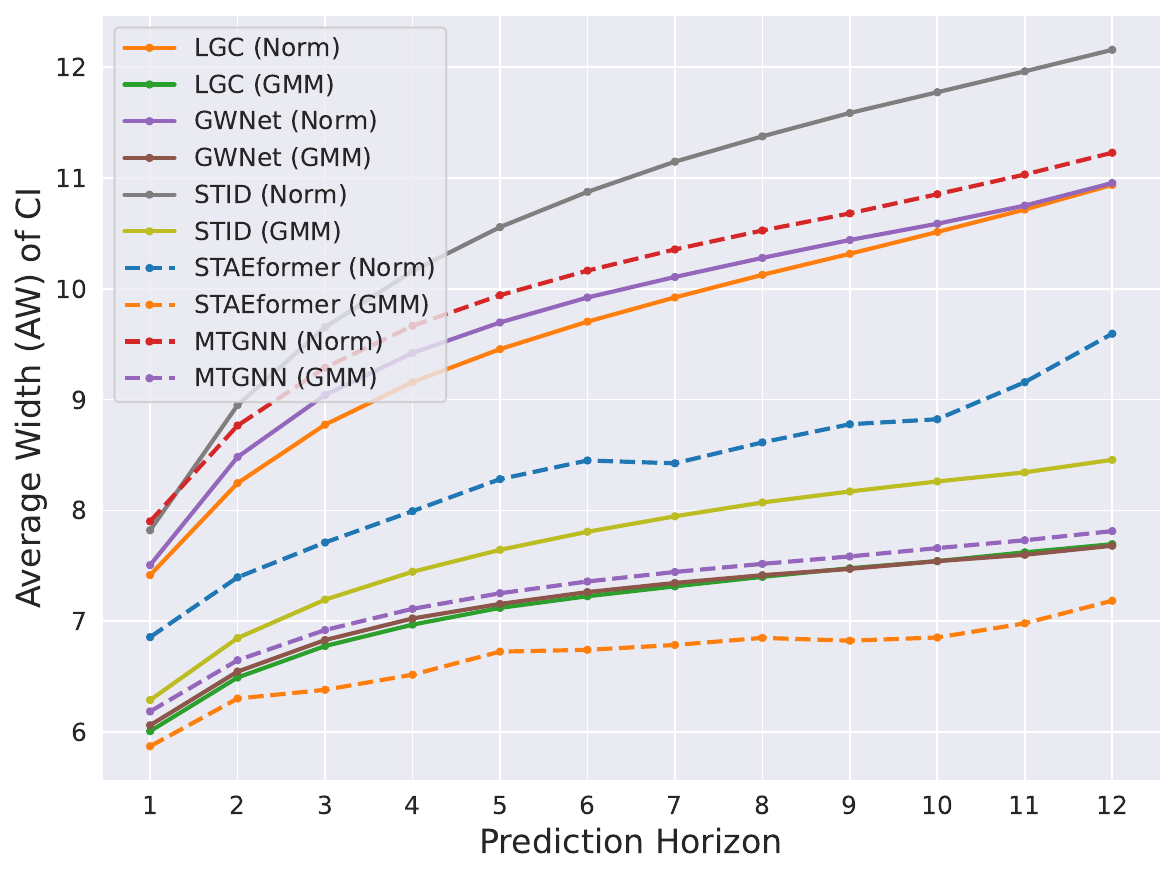}}
    \hfill
    \subfloat[PEMS-Bay]{\includegraphics[width=0.32\textwidth]{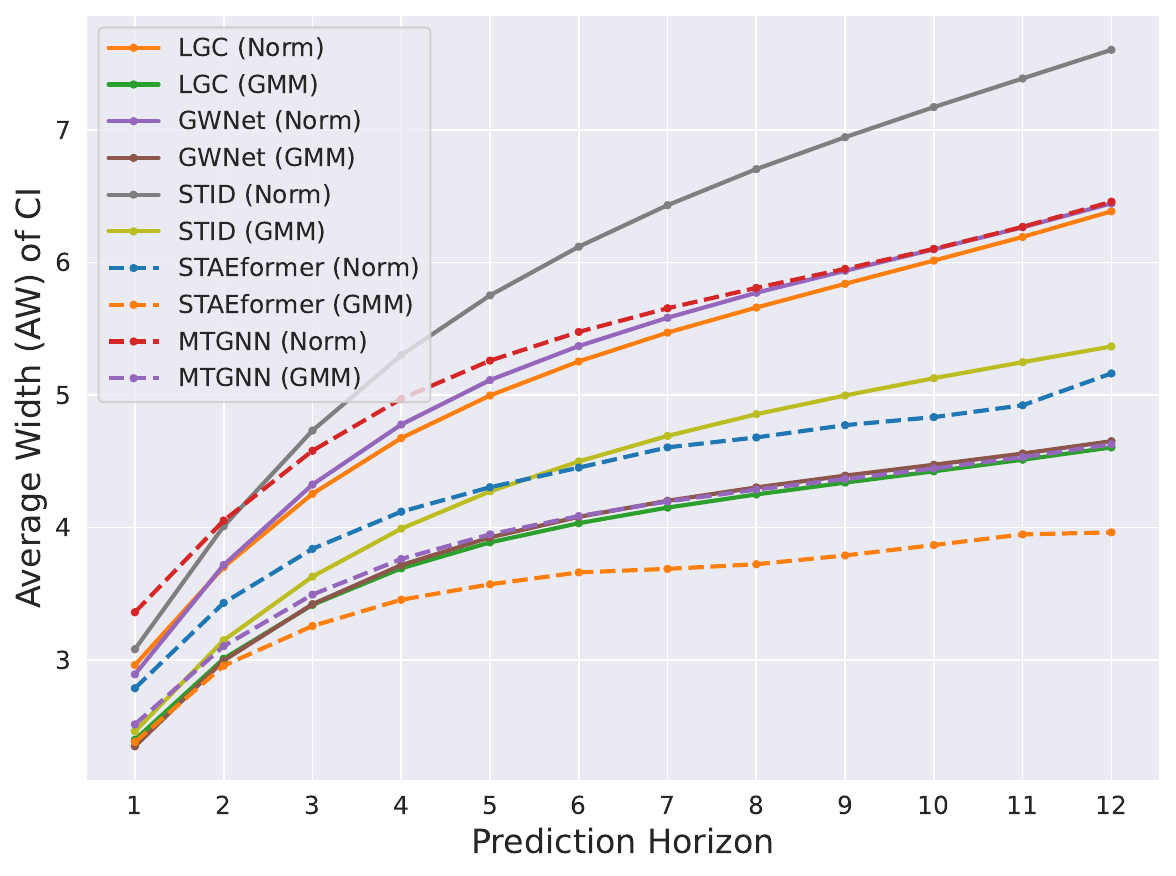}}
    \hfill
    \subfloat[SimBarcaSpd]{\includegraphics[width=0.32\textwidth]{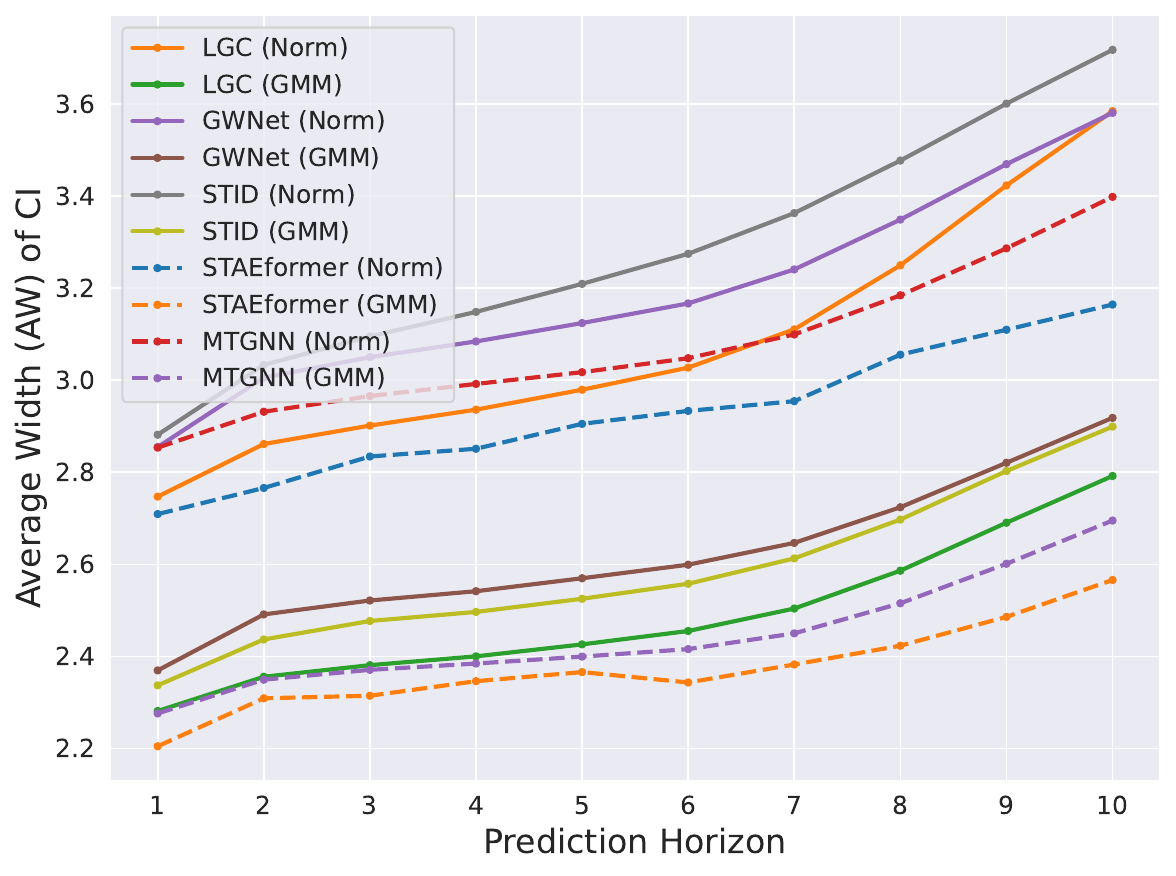}}
    \caption{mAW score of all probabilistic methods at all prediction horizons on three datasets. }
    \label{fig:maw_scores_by_prediction_horizon}
\end{figure*}

Figure~\ref{fig:maw_scores_by_prediction_horizon} shows the mAW scores of all probabilistic methods, which indicate the level of predicted uncertainty.
As expected, the predicted uncertainty also generally increases with the prediction horizon, since longer-horizon predictions are more challenging.
Notably, compared to the Normal variants, the GMM variants consistently have considerably lower mAW scores, indicating that the predictions are more informative with narrower confidence intervals.

\begin{figure*}[ht]
    \centering
    \subfloat[METR-LA]{\includegraphics[width=0.32\textwidth]{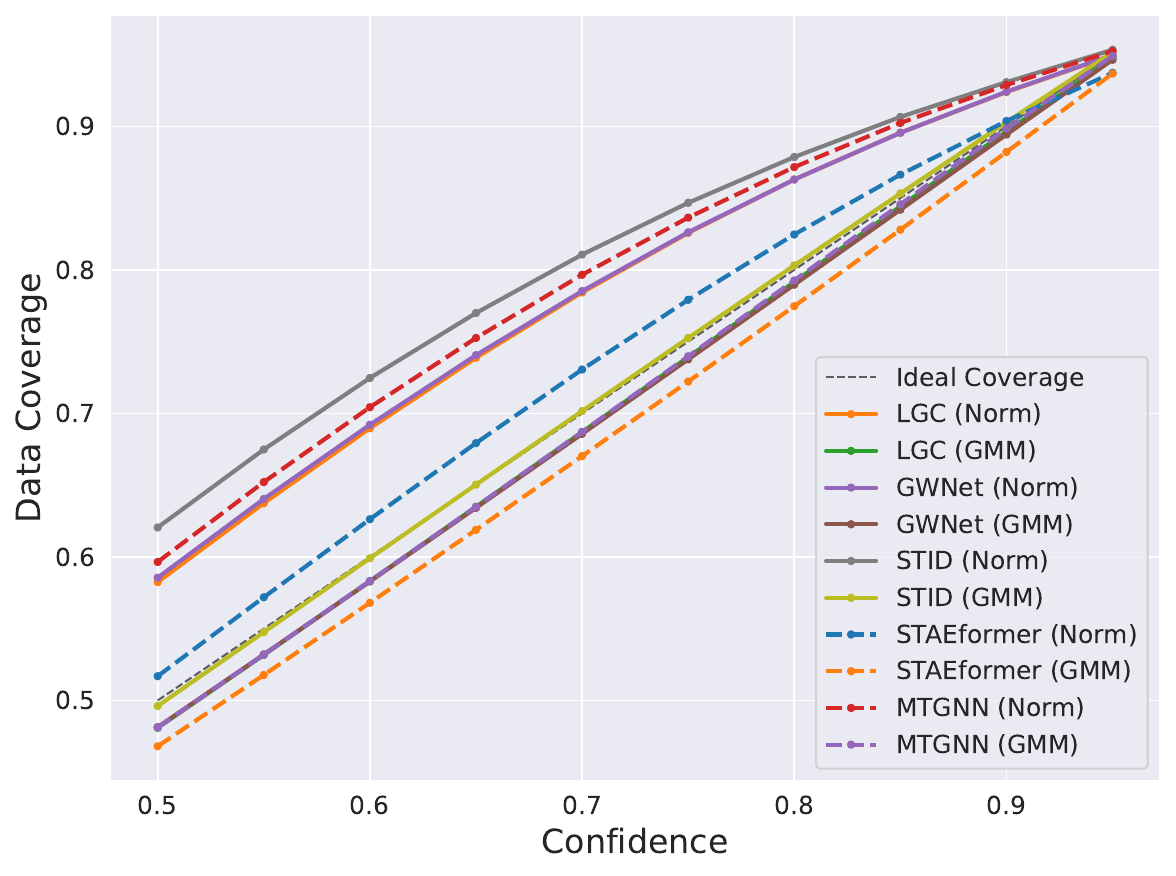}}
    \hfill
    \subfloat[PEMS-Bay]{\includegraphics[width=0.32\textwidth]{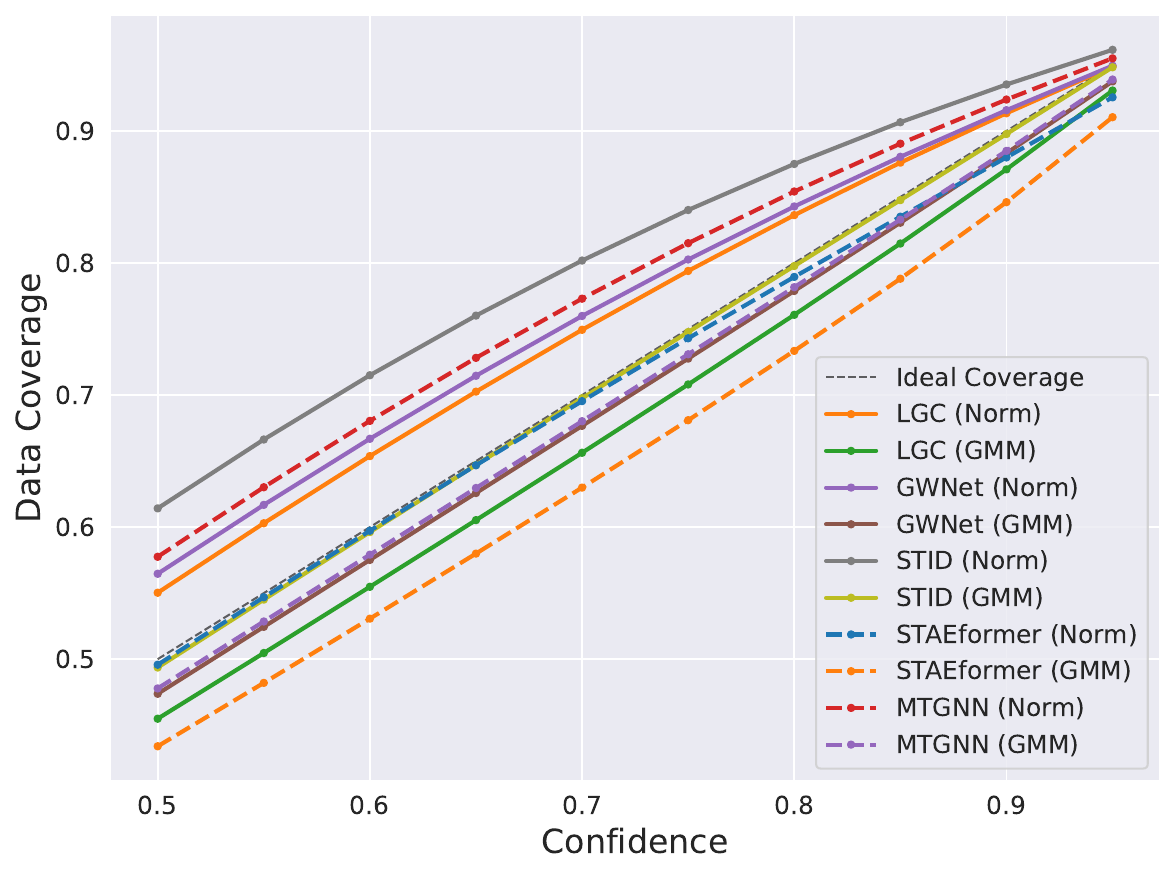}}
    \hfill
    \subfloat[SimBarcaSpd]{\includegraphics[width=0.32\textwidth]{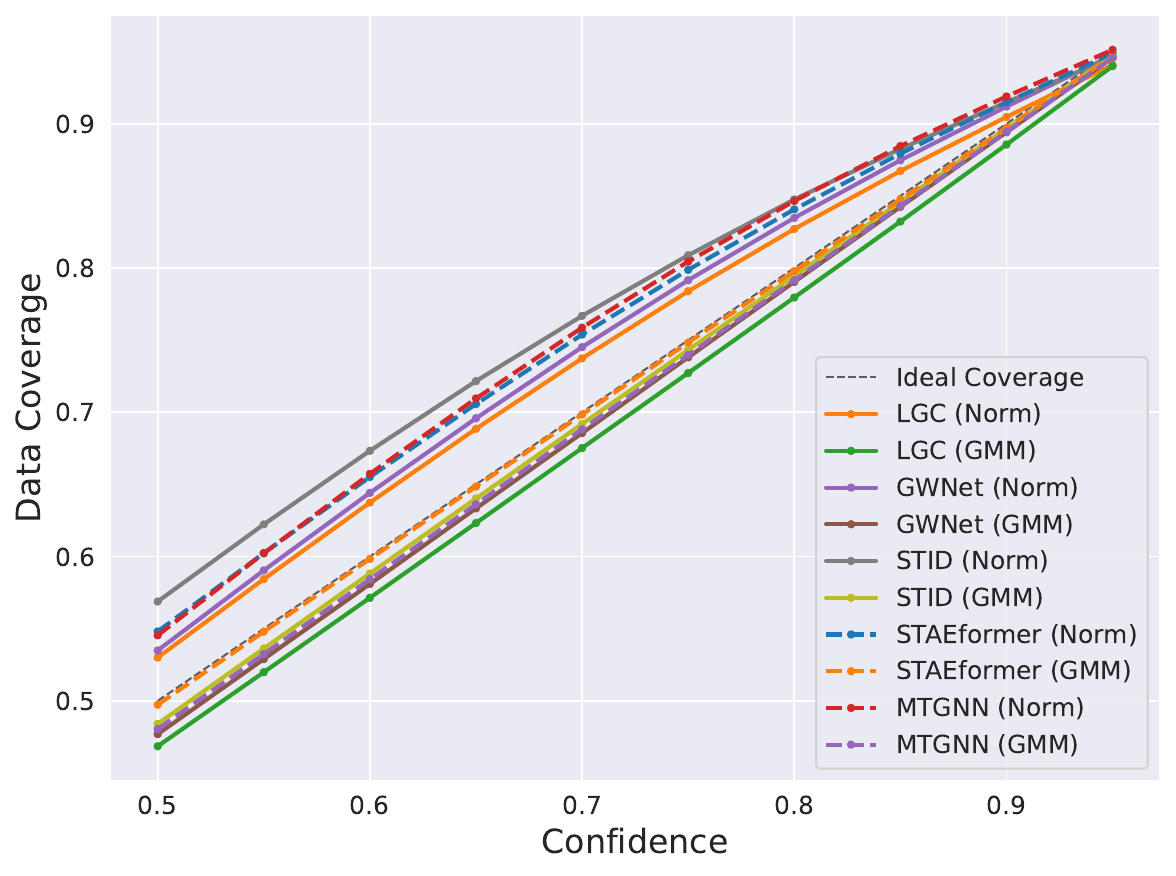}}
    \caption{Calibration curve of all probabilistic methods at different confidence interval levels on three datasets. }
    \label{fig:data_coverage_by_ci_level}
\end{figure*}

Figure~\ref{fig:data_coverage_by_ci_level} shows the calibration curves, i.e., the actual data coverage at different confidence levels.
The difference between the curves and the diagonal line indicates the calibration quality, and is quantified by the mCCE metric in Equation~\ref{equ: mcce}.
The calibration curve provides more details than the overall mCCE metric.
In the ideal case, the actual data coverage should be perfectly aligned with the confidence levels.
Any point above the ideal coverage line is under-confident, because the intervals cover more data than expected.
On the contrary, any point below the ideal line is over-confident.
Comparing across three datasets, complex backbone models generally have lower data coverage than simpler models.
For example, STAEformer-based GMM and Normal variants have lower data coverage than corresponding STID-based variants on all three datasets.
Meanwhile, GMM variants generally have lower data coverage than Normal variants of the same backbone architecture, indicating that the GMM confidence intervals have a more selective behavior.

\modified{Across the three evaluated datasets, the GMM variants of nearly all models are equal to or closer to the ideal calibration than Normal variants, which is also indicated by the mCCE metric in Table~\ref{tab:probabilistic_performance_metrics}.
Besides, most models tend to be consistently over-confident or consistently under-confident, i.e., the calibration curves of most models often lie on one side of the ideal line with few observed intersections.
For example, STID (Norm) is under-confident at all confidence levels for all three datasets, whereas LGC (GMM) is always over-confident.
}

\begin{figure*}[ht]
    \centering
    \subfloat[METR-LA]{\includegraphics[width=0.32\textwidth]{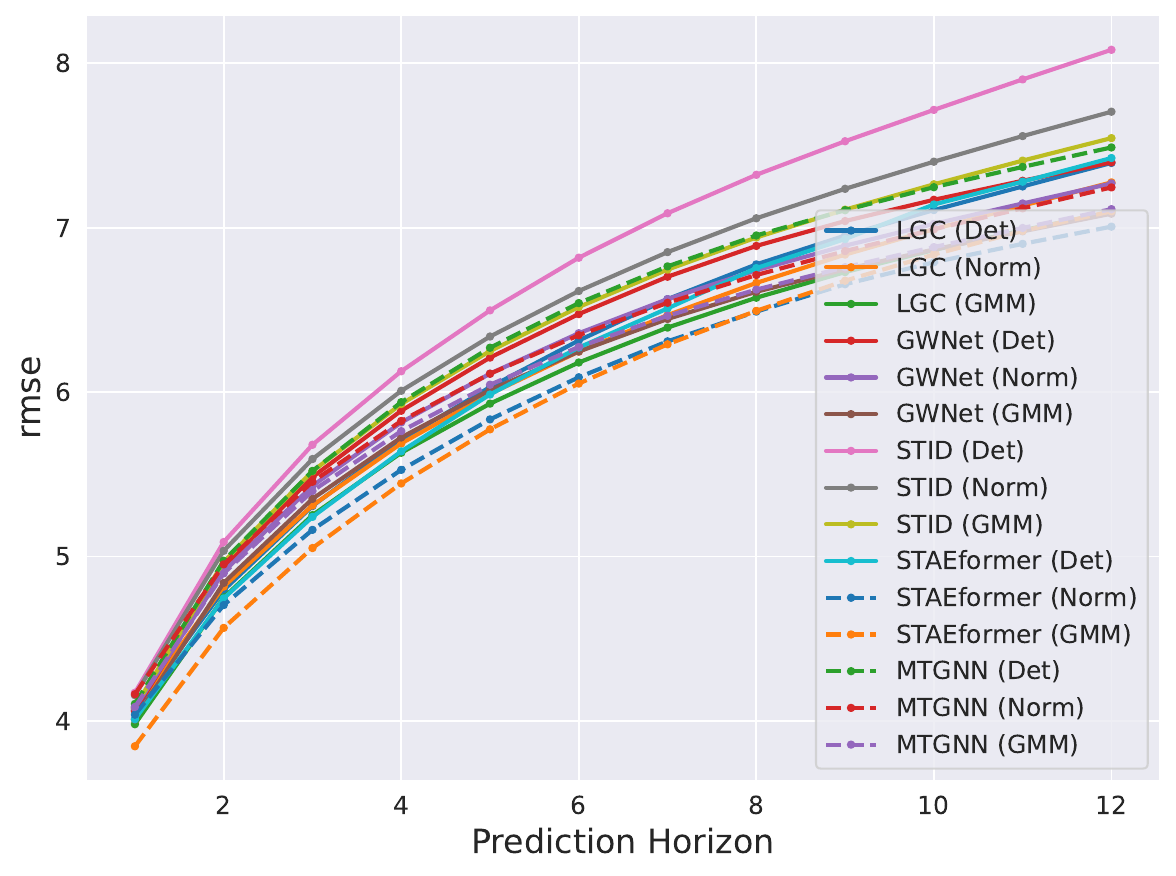}}
    \hfill
    \subfloat[PEMS-Bay]{\includegraphics[width=0.32\textwidth]{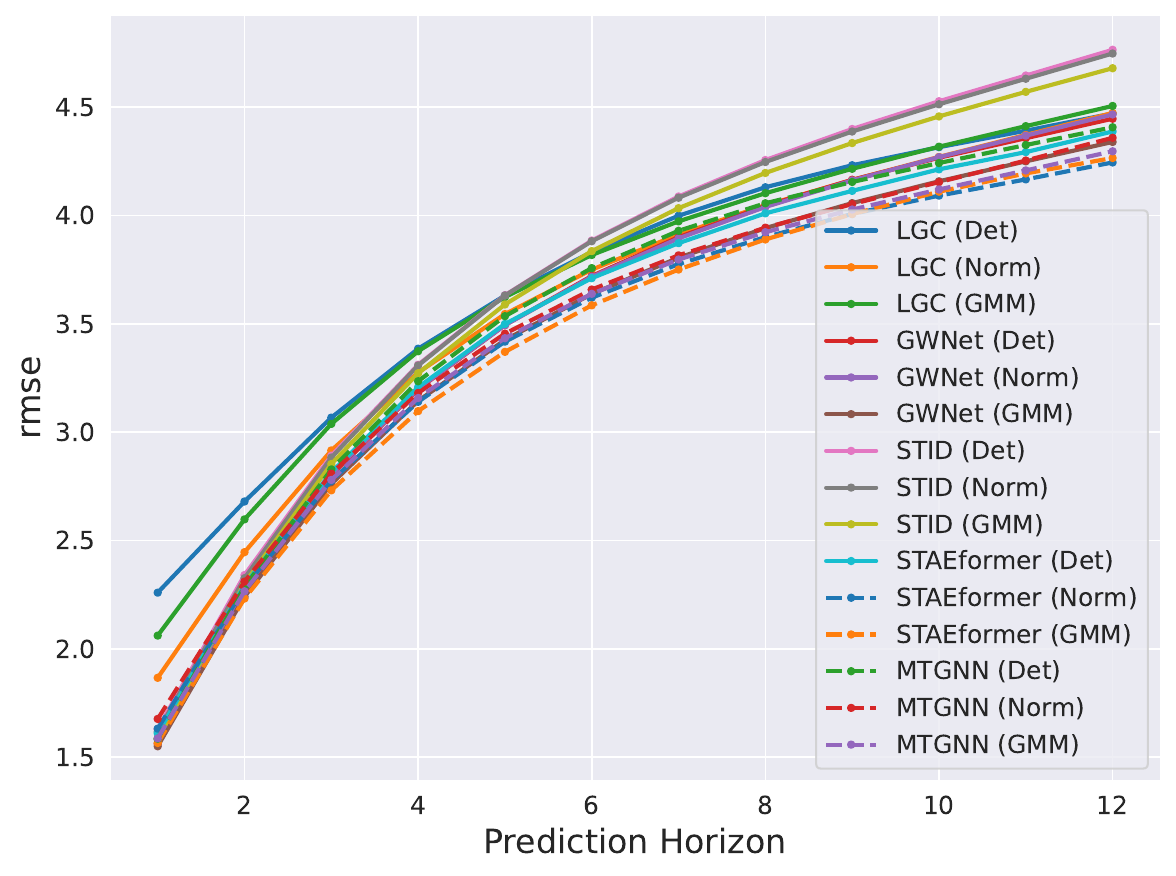}}
    \hfill
    \subfloat[SimBarcaSpd]{\includegraphics[width=0.32\textwidth]{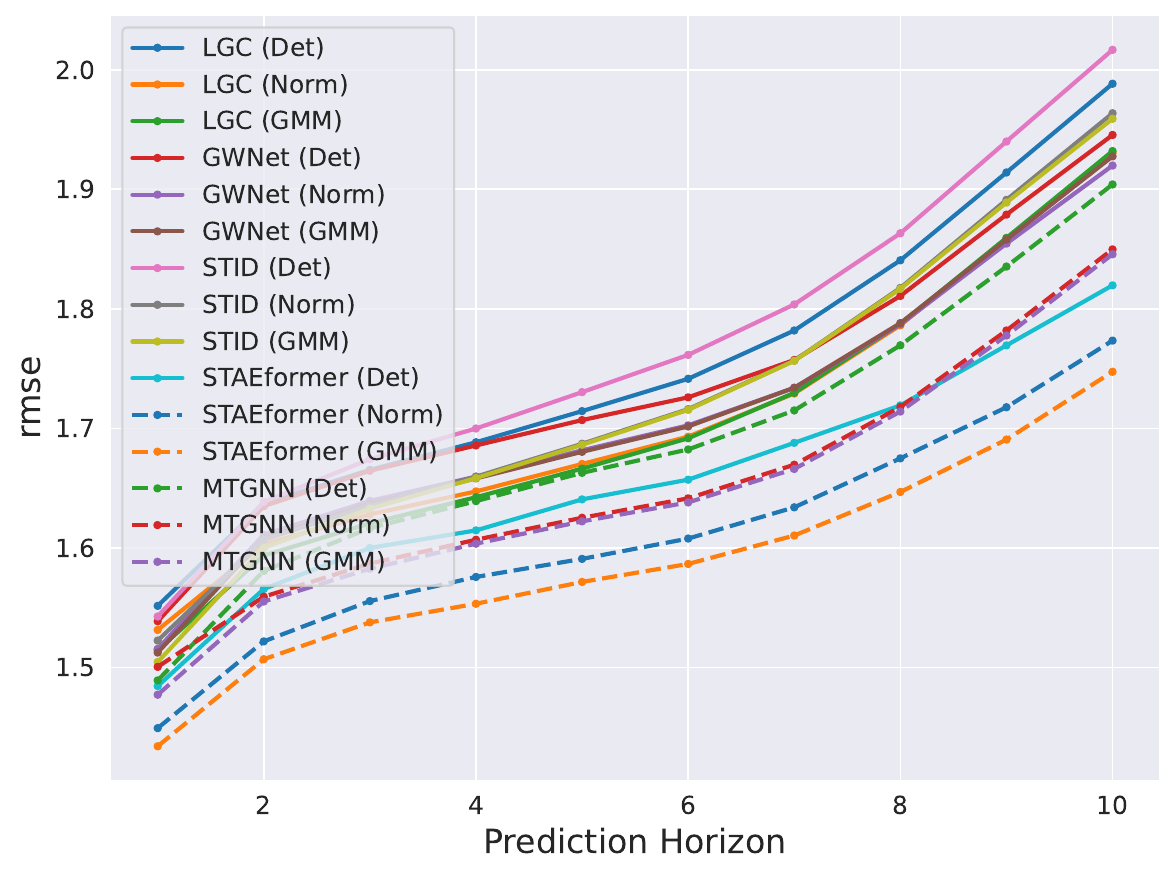}}
    \caption{RMSE of all methods at all prediction horizons on three datasets. }
    \label{fig:rmse_scores_by_prediction_horizon}
\end{figure*}

Figure~\ref{fig:rmse_scores_by_prediction_horizon} shows the RMSE of all methods at all prediction horizons, which increases with the prediction horizon as expected.
Still, the shape of the curves in SimBarcaSpd differs from the other two datasets, where RMSE increases more rapidly at the last few horizons.
Unlike the highway traffic in METR-LA and PEMS-Bay, SimBarcaSpd comes from an urban scenario with dense connections between road segments, and predicting longer horizons therefore becomes increasingly challenging due to complex interactions.

\subsection{\modified{Uncertainty Prediction Baselines}}
\label{subsec: uncertainty prediction baselines}
We compare the proposed GMM adaptation approach with the following representative baselines in uncertainty prediction.
To keep the scope concrete, we conduct experiments with the GWNet backbone on the METR-LA dataset.
All other experiment settings (e.g., optimizer, learning rate, training epochs, batch size) are aligned with previous experiments, and the results are shown in Table~\ref{tab:gwnet_uncertainty_results}.
For a fair comparison, the CRPS and confidence intervals of the baselines are derived from their natural prediction outputs, instead of forcing an empirical CDF to fit Equation~\ref{equ: crps} (details are provided in~\ref{app: metrics for probabilistic baselines}).

\begin{itemize}
    \item Monte Carlo Dropout \citep{gal2016dropout}. We activate the last dropout layer during inference, and perform 30 stochastic forward passes to obtain an empirical distribution.
    \item Quantile Regression \citep{koenker2001quantile}. We adapt the model to predict the quantiles using the pinball loss function.
    \item Conformal Prediction \citep{shafer2008conformal}. We derive an empirical error distribution with a held-out validation set, and derive confidence intervals from quantiles of the error distribution.
\end{itemize}

Among all uncertainty prediction methods, GMM has the best CRPS with a 0.19 gain compared to the second-best Quantile Regression, which clearly shows its advantage in probabilistic modeling for traffic forecasting.
The MC Dropout model appears to have deterministic behavior, as its predictive distribution has a very narrow range (1.06 mAW) and very high calibration error (0.603 mCCE).
Without considering this outlier, GMM delivers the most informative predictive intervals (7.19 mAW) and shares the best calibration error (0.010 mCCE) with Quantile prediction, which means GMM provides favorable probabilistic predictions.

\begin{table}[ht]
\centering
\caption{Uncertainty prediction benchmark. Lower is better for all metrics.}
\label{tab:gwnet_uncertainty_results}
\setlength{\tabcolsep}{1mm}
\begin{tabular}{lcccccc}
\toprule
Method & MAE & MAPE & RMSE & CRPS & mCCE & mAW \\
\midrule
GMM & 3.28 & 9.12\% & 6.08 & \textbf{2.26} & \textbf{0.010} & 7.19 \\
MC Dropout & \textbf{3.12} & 8.78\% & 6.26 & 2.89 & 0.603 & \textbf{1.06} \\
Quantile & 3.20 & 8.87\% & \textbf{6.05} & 2.45 & \textbf{0.010} & 8.56 \\
Conformal & \textbf{3.12} & \textbf{8.72\%} & 6.27 & 2.52 & 0.026 & 7.76 \\
\bottomrule
\end{tabular}
\end{table}

\subsection{\modified{Sensitivity and Efficiency}}
\label{subsec: sensitivity and efficiency}
To show the practicality of our GMM adaptation approach, we provide a sensitivity analysis of the GMM component number $K$ and profile the computational efficiency.
Similar to Section~\ref{subsec: uncertainty prediction baselines}, the sensitivity study uses the GWNet base model on METR-LA.
Table~\ref{tab:gmm anchor ablation} presents the evaluation metrics of a GMM-based model while varying the number of Gaussian components from 1 to 9.
The results show that the single-component (Gaussian) model has significantly worse CRPS than all others, which indicates the necessity of multi-modal probabilistic prediction in complex traffic forecasting tasks.
The models with multiple components have similar CRPS, and mAW improves with more components, but mCCE can degrade with overly complicated modeling.
For a balance of performance and complexity, we have defaulted to 5 components.

\begin{table}[ht]
\centering
\caption{Effect of GMM component number $K$. Lower is better for all metrics.}
\label{tab:gmm anchor ablation}
\setlength{\tabcolsep}{1mm}
\begin{tabular}{ccccccc}
\toprule
~$K$~ & MAE & MAPE & RMSE & CRPS & mCCE & mAW \\
\midrule
1 & 3.38 & 9.42\% & 6.17 & 2.52 & 0.081 & 9.67 \\
3 & 3.28 & 9.10\% & 6.09 & 2.26 & \textbf{0.009} & 7.27 \\
5 & 3.28 & 9.12\% & 6.08 & 2.26 & 0.010 & 7.19 \\
7 & \textbf{3.26} & 9.08\% & \textbf{6.07} & \textbf{2.25} & 0.013 & 7.11 \\
9 & 3.26 & \textbf{9.06\%} & 6.08 & 2.26 & 0.014 & \textbf{7.07} \\
\bottomrule
\end{tabular}
\end{table}

Theoretically, for a base model whose last layer (assumed to be a Linear layer) predicts $T$ future traffic values using a $D$-dimensional hidden space, its head has $TD$ parameters (omitting the bias for simplicity).
The GMM head for such a base model has a projection layer with $D^2$ parameters and $K$ Gaussian components, whose weight, mean and variance branches each have $T D$ parameters.
As a result, the GMM head will have $D^2 + 3KTD$ parameters.

Although GMM models have more parameters compared to the deterministic base models, the increase in computational cost is not significant. 
Table~\ref{tab:gmm profile configs} summarizes the step time and memory for each GMM model and the base model at the training and inference stages, respectively.
Concretely, a training step includes a forward pass, backpropagation, and an optimizer step, whereas an inference step is just the forward pass.
GPU memory is measured with batch size 32, and the step time is averaged over 100 repeats.
In general, a GMM model requires $\sim$2 ms more time per training step compared to its deterministic baseline, and the overhead is even smaller at inference time. 
The memory consumption of GMM is also slightly higher than that of the base model, because our adaptation only applies to the output layer, not the spatio-temporal backbone performing the majority of computation.
Therefore, the proposed GMM adaptation is a promising approach that brings substantial performance gain in probabilistic metrics and only requires slightly more resources than a deterministic base model.

\begin{table}[ht]
\centering
\caption{GMM overhead on METR-LA dataset, with the deterministic model in $(\cdot)$. Performance measured on a single Nvidia RTX4090.}
\label{tab:gmm profile configs}
\setlength{\tabcolsep}{1mm}
\resizebox{\columnwidth}{!}{%
\begin{tabular}{lrr|rr}
\toprule
Method & \multicolumn{2}{c}{Time (ms/step)} & \multicolumn{2}{c}{GPU Mem. (MB)} \\
\cmidrule(lr){2-3} \cmidrule(lr){4-5}
& Training & Inference & Training & Inference \\
\midrule
STID & 3.0 (1.6) & 0.8 (0.5) & 68 (37) & 32 (27) \\
LGC & 7.5 (6.9) & 3.1 (3.0) & 973 (973) & 565 (564) \\
GWNet & 14.7 (13.2) & 3.9 (3.7) & 585 (525) & 281 (277) \\
MTGNN & 11.5 (11.0) & 3.1 (3.0) & 346 (316) & 147 (147) \\
STAE & 67.7 (66.1) & 21.5 (21.3) & 3989 (3982) & 707 (702) \\
\bottomrule
\end{tabular}
}
\end{table}

\subsection{Prediction Examples}
\label{subsec: prediction examples}
\begin{figure*}[ht]
    \centering
    \subfloat[Segment 113]{\includegraphics[width=0.32\textwidth]{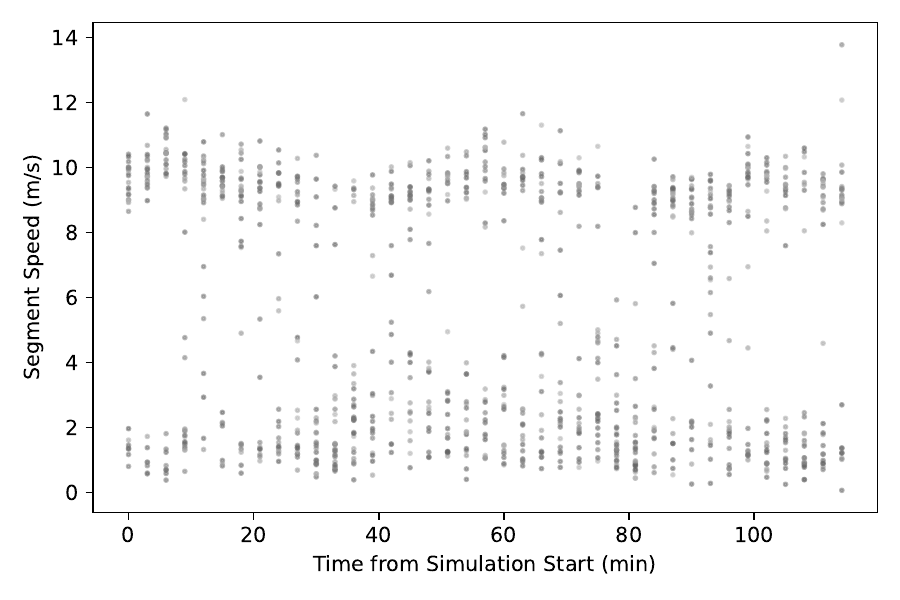}}
    \hfill
    \subfloat[Segment 345]{\includegraphics[width=0.32\textwidth]{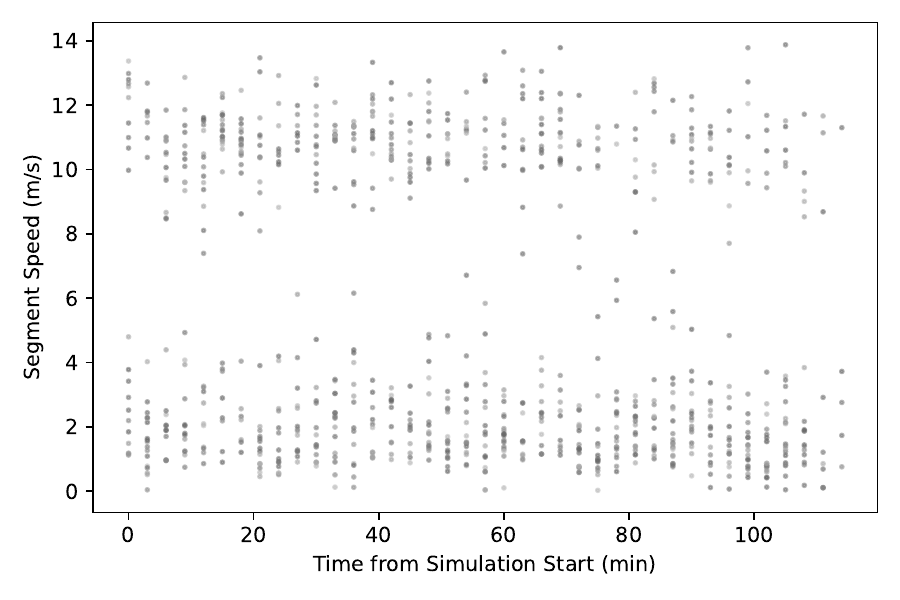}}
    \hfill
    \subfloat[Segment 1068]{\includegraphics[width=0.32\textwidth]{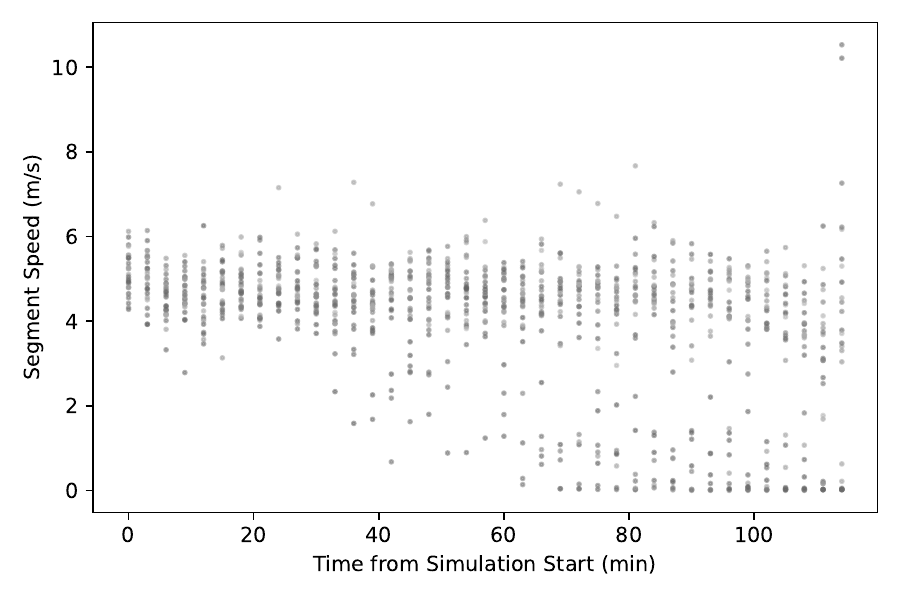}}
    \caption{Test-set data distribution in SimBarcaSpd. The data from 26 test-set sessions are plotted according to their simulation time. The scatter plot clearly shows that the traffic speed at these segments has multiple modes, e.g., a congested mode and a free-flow mode.}
    \label{fig:multi-modal distribution of data}
\end{figure*}

\begin{figure*}[ht]
    \centering
    \subfloat[Segment 113]{\includegraphics[width=0.32\textwidth]{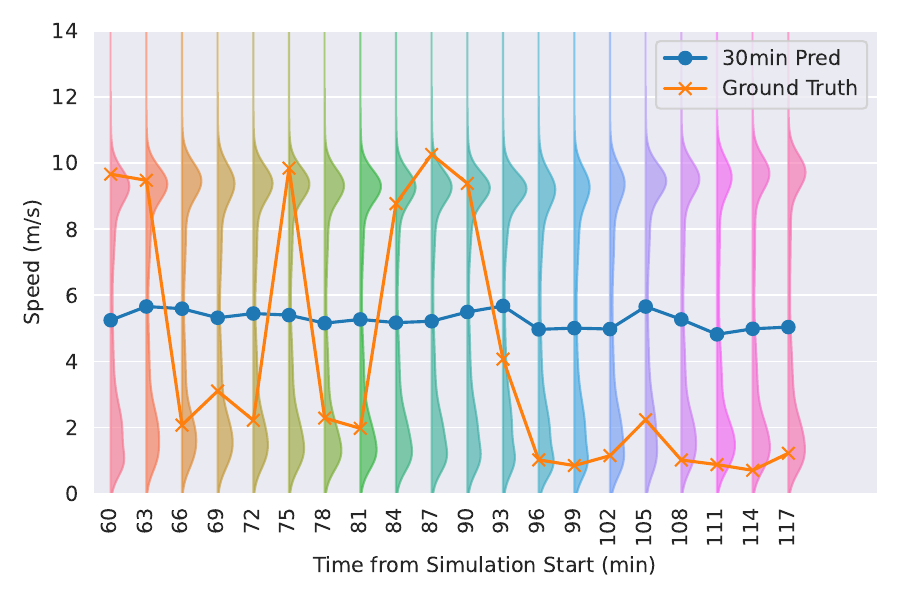}}
    \hfill
    \subfloat[Segment 345]{\includegraphics[width=0.32\textwidth]{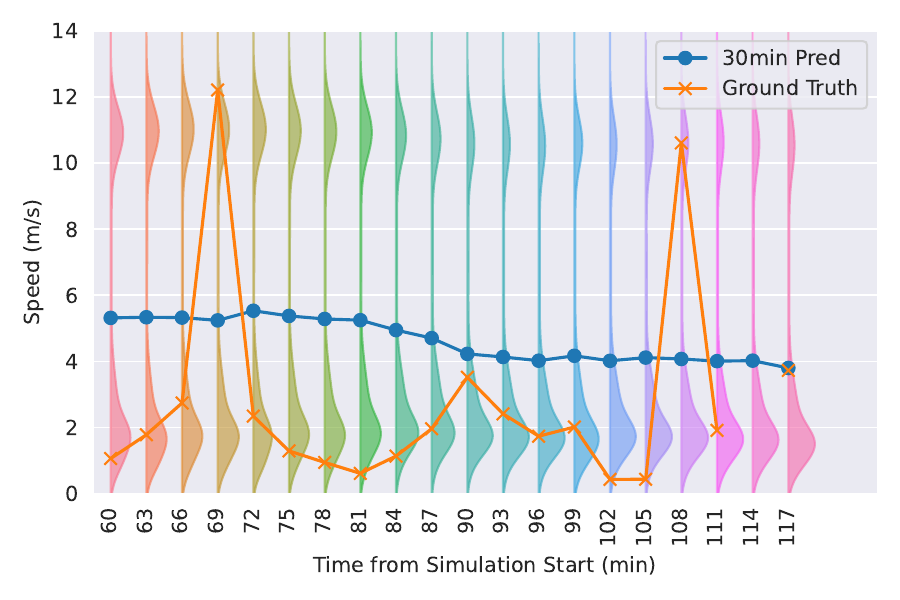}}
    \hfill
    \subfloat[Segment 1068]{\includegraphics[width=0.32\textwidth]{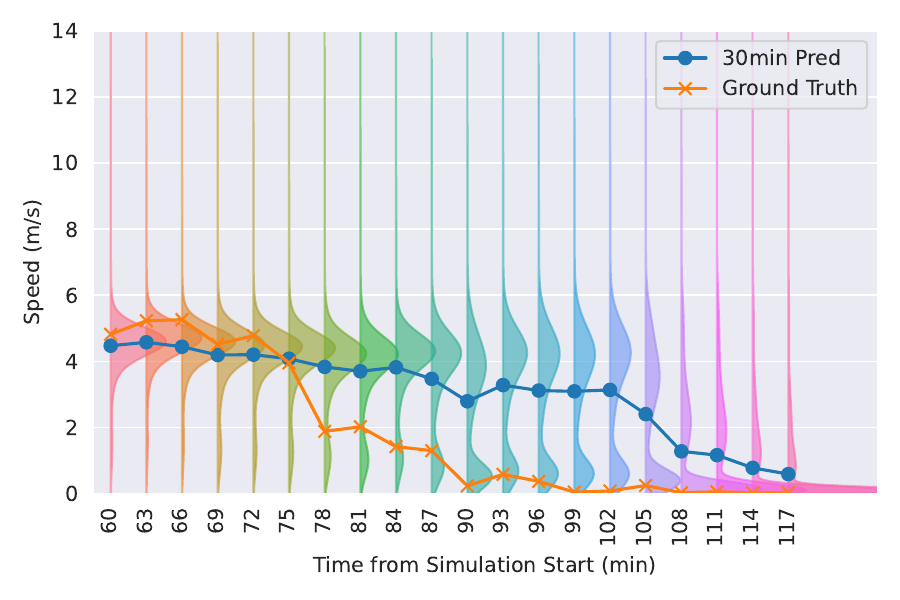}}
    \caption{The 30-minute-ahead predictions (10 steps) of LGC (GMM) for a high-traffic-demand scenario in SimBarcaSpd. \modified{We use 30-minute input window, and thus the first 30-minute-ahead prediction corresponds to 60 min. The density ridges are colored by time step for easier visual identification of per-step predictions.}}
    \label{fig:multi-modal predictions}
\end{figure*}

To further illustrate the challenge of modeling urban traffic, Figure~\ref{fig:multi-modal distribution of data} shows the test-set data distribution for three representative road segments in SimBarcaSpd.
From these visualizations, it is clear that the traffic speed for urban roads can have multiple modes (e.g., congested vs free-flow), while the middle range is less likely.
Figure~\ref{fig:multi-modal predictions} presents the 30-minute-ahead prediction examples of the LGC-GMM at these segments; data from a high-traffic-demand scenario are selected for visualization.
In both segments 113 and 345, the deterministic prediction is a stable-looking curve around a middle value, yet road speeds in reality flip between low and high values.
Under such scenarios, deterministic predictions are very misleading, but this is a fundamental limitation coming from the training paradigm, where the model is forced to predict a single value to minimize the absolute error.
In contrast, our GMM adaptation allows the model to produce multi-modal distributions, and it is clear that for both segments, the GMM predictions have two distinct modes that can cover the two possibilities in reality.
For segment 345, the model can predict a high-speed mode around 11 m/s and a low-speed one around 2 m/s, both of which align with the scatter points in the data distribution shown in Figure~\ref{fig:multi-modal distribution of data}.
Another advantage of multi-modal predictions is faster adaptation to changes.
For segment 1068, traffic congestion builds up from 75 min, and the deterministic prediction reacts slowly, while the mode for the congested regime appears in the GMM prediction as early as 78 min.
This ability can be crucial in traffic management, where earlier warnings of congestion allow more effective mitigation measures.

\begin{figure*}
    \centering
    \includegraphics[width=\linewidth]{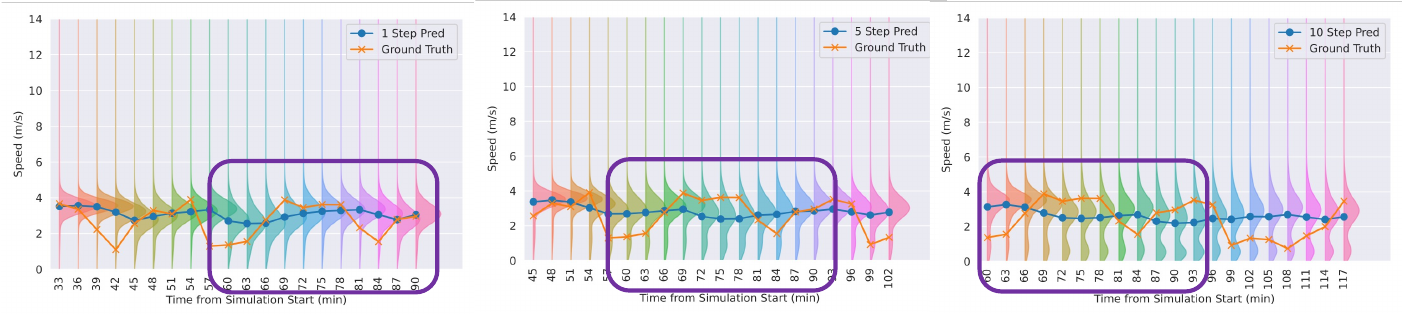}
    \caption{One-, five-, and ten-step predictions of LGC (GMM) for road segment 806 in SimBarcaSpd. The density ridges are colored for better visualization.}
    \label{fig:uncertainty prediction at horizons}
\end{figure*}

Figure~\ref{fig:uncertainty prediction at horizons} shows how the prediction time horizon affects the predicted distributions.
For the same simulation time (e.g., 60 min to 90 min), the one-step-ahead prediction has a sharper single-mode distribution, since predicting the immediate next step is relatively easier.
As the horizon extends to five and ten steps, the predicted distribution becomes more uncertain for the same time span, with a wider spread and multiple modes.
This behavior shows the GMM model's ability to capture the increasing uncertainty with longer future horizons.

\subsection{Robustness to Data Quality}
\label{subsec: impact of data quality}
To investigate the impact of data quality on probabilistic traffic forecasting, we conduct an additional experiment using the SimBarca dataset \citep{xiong2025multi}, featuring drone-measured high-frequency traffic speed data collected every 5 seconds.
Concretely, SimBarca and our derived SimBarcaSpd dataset have the same input and output window (30 min), but SimBarca has finer temporal resolution (5 seconds), providing more detailed traffic dynamics than SimBarcaSpd (3 minutes).
We adapt the LGC model to SimBarca by adding a temporal convolution layer before the LSTM to downsample the higher-resolution input, while keeping the rest of the architecture unchanged.

We study the impact of data quality from two aspects:
\begin{itemize}
    \item Partial coverage: when sensors cover only a random 10\% of the road segments \citep{xiong2025multi}.
    \item Temporal detail: when input data are provided every 5 seconds (more details) or 3 minutes (marked as Low Res).
\end{itemize}

\begin{figure}[ht]
    \centering
    \includegraphics[width=0.8\linewidth]{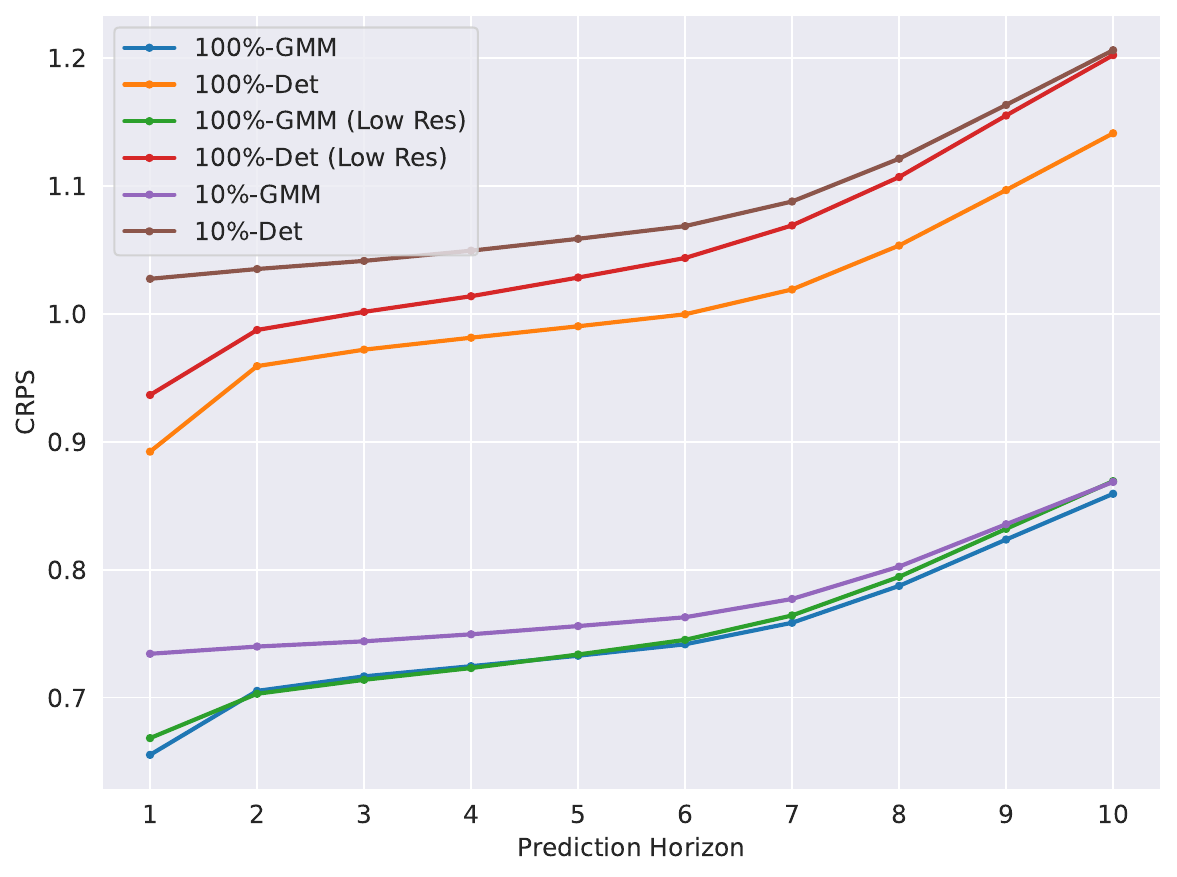}
    \caption{CRPS scores of LGC variants on SimBarca under different data quality settings.}
    \label{fig:simbarca metrics}
\end{figure}

Figure~\ref{fig:simbarca metrics} shows the CRPS scores of the LGC model on SimBarcaSpd (marked as Low Res) and the adapted LGC model on SimBarca under different data quality settings.
Table~\ref{tab:avg_crps_mae_across_horizons} also summarizes the average CRPS across all prediction horizons for quantitative comparison.
The percentage in parentheses indicates the increased CRPS compared to the ideal case.

The ideal case is when the model has access to high-resolution data with full coverage, and both models perform best relative to degraded data-quality settings.
When only 10\% of the road segments are covered, the CRPS for the probabilistic GMM method increases by 3.5\%, while the deterministic Det method has a larger CRPS increase of 7.5\%.
This observation is also consistent with the larger gap in Figure~\ref{fig:simbarca metrics} between the ideal and 10\% coverage curves for Det compared to GMM.
The impact of lower temporal resolution is less significant than coverage, where the CRPS for GMM increases by only 0.6\% and Det increases by 4.3\%.
These comparisons suggest that multi-modal probabilistic modeling can provide more accurate and informative predictions while also being more robust to data-quality issues such as sensor coverage and data frequency.

\begin{table}[ht]
\centering
\caption{Average CRPS with different input data quality. Numbers in parentheses indicate the percentage increase compared to the ideal case, lower values are preferred.}
\label{tab:avg_crps_mae_across_horizons}
\setlength{\tabcolsep}{1mm}
\begin{tabular}{lccc}
\toprule
Model & Ideal & 10\% coverage & Low resolution\\
\midrule
GMM & 0.75 & 0.78 (+3.50\%) & 0.75 (+0.60\%) \\
Det & 1.01 & 1.09 (+7.50\%) & 1.05 (+4.30\%) \\
\bottomrule
\end{tabular}
\end{table}

\section{Conclusions}
\label{sec: conclusions}

In summary, this paper presents a universal approach to transform existing spatio-temporal forecasting models into probabilistic ones by replacing only the final output layer with a novel GMM layer.
Our adaptation method is simple, effective, and can be applied to a wide range of backbone architectures with minimal modifications to the model structure and training procedure.
A comprehensive evaluation methodology has been introduced to assess the quality of predictive distributions, through which we show that probabilistic modeling consistently improves CRPS over deterministic approaches across multiple datasets and architectures.
Compared to deterministic predictions, which are often vulnerable to false-confidence issues, our multi-modal probabilistic predictions can better capture the complex and multi-modal nature of traffic data, providing more informative and safer predictions for real-world applications.
Additional analyses have demonstrated that a probabilistic model equipped with the GMM prediction layer can be more robust to data quality issues such as coverage (missing rate) and temporal resolution.

\modified{However, the additional modeling flexibility of GMM may lead to worse calibration in a few cases, e.g., STAEformer on PEMS-Bay, despite having better CRPS.
On deterministic evaluations, GMM-based models tend to have better L2 error (RMSE) in most settings, while the L1 error metrics are often slightly worse than native deterministic models that are trained directly with L1 loss.
Therefore, we believe our method is more suitable for traffic scenarios with complex state changes, e.g., in the busy urban road networks of SimBarcaSpd.
Another limitation of this work is that we only model the marginal distribution, and we believe learning the full distribution with covariances will be a promising future direction.}

We hope our work can raise awareness of probabilistic modeling in the community and provide a novel perspective for evaluating spatio-temporal forecasting models.
Future research could explore the potential of multi-modal distributions with more advanced base architectures and probabilistic modeling techniques.

\section*{CRediT Author Contributions}
\label{sec: author contributions}
Weijiang Xiong: Conceptualization, Methodology, Data Curation, Software, Writing - Original Draft. 
Robert Fonod: Conceptualization, Methodology, Writing - Review \& Editing, Supervision.
Nikolas Geroliminis: Conceptualization, Methodology, Writing - Review \& Editing, Supervision, Funding Acquisition.

\section*{Acknowledgements}
This work is supported by the Swiss National Science Foundation (SNSF) under NCCR Automation, grant agreement number 51NF40\_180545.

\bibliographystyle{elsarticle-harv}
\bibliography{references}

\appendix

\section{Reference Values and Scales}
\label{app: reference values and scales}

In this section, we provide more details to justify the choices of reference values and scales in Section~\ref{subsec: gmm layer}.
In PyTorch \citep{paszke2019pytorch}, the weights and biases of a linear layer are both initialized with a uniform distribution $\mathcal{U}(-\sqrt{k}, \sqrt{k})$, where $k$ is the inverse of the input feature dimension, i.e., $k=1/C_{in}$.
We consider a vector $\mathbf{w} \in R^{C_{in}}$ in the weight matrix, and an input data vector $\mathbf{x} \in R^{C_{in}}$.
The variance of their dot product is:
\begin{equation}
    \text{Var}(\mathbf{w}^T\mathbf{x}) =
    \sum_1^{C_{in}}\text{Var}(w_i * x_i).
\end{equation}
When $\mathbf{x}$ is normalized to have zero mean and unit variance,
\begin{equation}
    \text{Var}(w_i * x_i) = 
    \text{Var}(w_i) * \text{Var}(x_i) = 
    \frac{k}{3} * 1 = \frac{k}{3}.
\end{equation}
Therefore,
\begin{equation}
    \text{Var}(\mathbf{w}^T\mathbf{x}) = \frac{k}{3} \times C_{in} = \frac{1}{3}.
\end{equation}
This means that, upon initialization, the output of a linear layer given Z-score-normalized inputs will likely lie in a small range around zero, because the biases also have zero mean. 
Therefore, we designed the mean value formulation in Section~\ref{subsec: gmm layer}, so that the ranges of the transformed mean values can jointly provide a good coverage of the data space even if the direct output of the linear layer stays within the unit range.

\section{Metrics for probabilistic baselines}
\label{app: metrics for probabilistic baselines}

In this section, we describe the procedure of computing point estimates (for deterministic metrics), CRPS and confidence intervals for the baseline methods involved in Section~\ref{subsec: uncertainty prediction baselines}.

MC Dropout is treated as an ensemble: each stochastic forward pass creates a sample, and all the samples together form an empirical distribution.
The point estimate is simply the average of the samples, the confidence intervals are derived from the quantiles of the samples.
For CRPS, we apply the formula of \citep{gneiting2007strictly}:
\begin{equation}
    \mathrm{CRPS}(F, y)
    =
    \mathbb{E}|X-y|
    -
    \frac{1}{2}\mathbb{E}|X-X'|, 
    \label{equ: crps alternative}
\end{equation}
where $y$ is the ground-truth value, $X$ is the predictive distribution represented by the samples, and $X'$ is an independent copy of the same random variable.

Quantile Regression directly predicts selected quantile levels, which are
\begin{equation}
\tau \in \{0.01, 0.05, 0.1, 0.2, 0.3, 0.4, 0.5, 0.6, 0.7, 0.8, 0.9, 0.95, 0.99\}
\end{equation}
Therefore, its confidence intervals can be naturally composed by selecting the upper and lower bounds, e.g., the 90\% confidence interval ranges from the 5\% quantile to the 95\% quantile.
We interpolate the predicted quantiles when the required quantile level is not present in the set.
Since the quantile-based prediction can be regarded as a numerical approximation of the predicted CDF, its mean estimate is the average of all quantiles:
\begin{equation}
\hat{y} 
= 
\sum_{k=1}^{K-1}
\frac{\tau_{k+1}-\tau_k}{2}
\left(q_{\tau_k}+q_{\tau_{k+1}}\right),
\end{equation}
where $\tau_k$ is the $k$-th quantile level, and $q_{\tau_k}$ is the predicted quantile value. 
Its CRPS is estimated by numerical integration of the corresponding pinball losses over the quantile grid following \citep{arnold2024decompositions}:
\begin{equation}
\operatorname{CRPS}(F,y)
=
2\int_0^1 \rho_\tau(y-q_\tau)\,d\tau,
\end{equation}
where the pinball loss for a given quantile $\tau$ is defined as
\begin{equation}
\rho_\tau(u)
=
u\left(\tau-\mathbf{1}\{u<0\}\right).
\end{equation}

Conformal Prediction starts from a deterministic prediction and builds its predictive distribution based on the prediction errors on a calibration set.
We group the calibration errors by location and time to obtain a sample-based residual distribution $R$ for each location and prediction horizon.
Since the underlying model is deterministic, the point prediction can be naturally obtained from the model.
In fact, the model's probabilistic predictions can be regarded as the combination of deterministic prediction and residual distribution:
\begin{equation}
    \hat{Y} = \hat{y} + R,
\end{equation}
where $\hat{y}$ is the deterministic prediction and $\hat{Y}$ is the probabilistic prediction.
The confidence intervals can be estimated by adding the quantiles of residual distribution $R$ to the deterministic prediction.
CRPS can also be computed using Equation~\ref{equ: crps alternative} with the samples in the residual distribution.

\section{Numerical Accuracy of Density Estimation}
\label{app: numerical accuracy of density estimation}

The GMM-based density functions are derived with numerical density estimation, which can affect the confidence intervals in Algorithm~\ref{alg: hdps derivation} and CRPS estimation with Equation~\ref{equ: crps}.
By default, we evaluate the density function on 2000 points evenly spaced from 0 to the maximum speed in Table~\ref{tab:dataset_specs}.
Table~\ref{tab:numerical-sensitivity} shows the probabilistic metrics when a GWNet-GMM model is evaluated on METR-LA dataset with different density resolutions.
The CRPS estimation is more robust to numerical accuracy, and an estimation with 250 points already has fair precision.
However, mCCE and mAW converge more slowly even if the number of evaluated points reaches 4000, due to the challenge of estimating the boundary of predicted intervals.
In our experiments, we choose 2000 points as the default evaluation configuration for a balanced numerical accuracy and computation efficiency.
Since the error in numerical density estimation cannot be fully avoided, we remind the readers that the default numerical resolution may underestimate mAW and result in a higher mCCE.
Therefore, a small difference in probabilistic metrics may not have enough significance, e.g., $\sim$0.002 CRPS, $\sim$0.005 mCCE or $\sim$0.05 mAW.

\begin{table}[ht]
\centering
\caption{Sensitivity of GMM evaluation metrics to the density resolution.}
\label{tab:numerical-sensitivity}
\resizebox{\columnwidth}{!}{%
\begin{tabular}{rrrrrrrr}
\toprule
Points & CRPS & 50\% Cov. & 50\% Width & 90\% Cov. & 90\% Width & mCCE & mAW \\
\midrule
100  & 2.283167 & 0.277243 & 2.549166 & 0.835138 &  9.835323 & 0.147609 & 6.151771 \\
250  & 2.263658 & 0.410371 & 3.199314 & 0.880081 & 10.498338 & 0.054790 & 6.811724 \\
500  & 2.260575 & 0.452867 & 3.416636 & 0.890719 & 10.717607 & 0.028066 & 7.030643 \\
1000 & 2.259528 & 0.472948 & 3.525114 & 0.895333 & 10.827016 & 0.015748 & 7.139887 \\
2000 & 2.259134 & 0.482663 & 3.579383 & 0.897481 & 10.881686 & 0.009851 & 7.194474 \\
4000 & 2.258909 & 0.487602 & 3.606475 & 0.898503 & 10.909010 & 0.006933 & 7.221750 \\
\bottomrule
\end{tabular}%
}
\end{table}

\end{document}